\documentclass{article}
\usepackage{arxiv}

\usepackage[utf8]{inputenc} 
\usepackage[T1]{fontenc}    
\usepackage{hyperref}       
\usepackage{url}            
\usepackage{booktabs}       
\usepackage{amsfonts}       
\usepackage{nicefrac}       
\usepackage{microtype}      
\usepackage{lipsum}		
\usepackage{graphicx}
\usepackage{doi}

\usepackage{amssymb}
\usepackage[figuresright]{rotating}

\usepackage{graphicx}
\usepackage{stackengine} 
\stackMath
\usepackage{float}
\usepackage{subcaption}
\usepackage{xspace}
\usepackage{amsmath}
\usepackage{makecell}
\usepackage{svg}
\usepackage{amsmath}
\usepackage{tikz}
\usepackage{paralist}
\usepackage{algorithmic}  
\usepackage{algorithm}
\usepackage[algo2e]{algorithm2e}
\usepackage{xkcdcolors}
\usetikzlibrary{shapes,arrows,chains,shadows,positioning}
\usetikzlibrary{arrows.meta}
\usetikzlibrary{decorations.pathreplacing}
\usepackage{geometry}
\usepackage{textcomp}
\usepackage{pgfplots}
\pgfplotsset{width=10cm,compat=1.9}

\usepackage[utf8]{inputenc}
\usepackage[T1]{fontenc}    
\usepackage{hyperref}       
\usepackage{url}            
\usepackage{multirow}
\usepackage{comment}
\DeclareUnicodeCharacter{2212}{-}

\title{Physics Informed Machine Learning for Chemistry Tabulation}

\author{
\hspace{1mm}Amol Salunkhe\\
 \hspace{1mm}University at Buffalo, 338 Davis Hall\\
  Buffalo, New York 14260\\
\And
\hspace{1mm}Dwyer Deighan\\
 \hspace{1mm}University at Buffalo, 338 Davis Hall\\
  Buffalo, New York 14260\\
 \And
  \hspace{1mm}Paul E. DesJardin\\
 \hspace{1mm}University at Buffalo, 338 Davis Hall\\
  Buffalo, New York 14260\\
\And
\hspace{1mm}Varun Chandola\\
 \hspace{1mm}University at Buffalo, 338 Davis Hall\\
  Buffalo, New York 14260\\
}

\renewcommand{\shorttitle}{\textit{Physics Informed} ML for Chemistry Tabulation}

\hypersetup{
pdftitle={Physics Informed Machine Learning for Chemistry Tabulation},
pdfsubject={q-bio.NC, q-bio.QM},
pdfauthor={Amol Salunkhe},
pdfkeywords={Physics Informed Machine Learning, Deep Neural Networks, Combustion, Fluid Dynamics, Chemistry Tabulation},
}

\begin{document}
\maketitle

\begin{abstract}
Modeling of turbulent combustion system requires modeling the underlying chemistry and the turbulent flow. Solving both systems simultaneously is computationally prohibitive. Instead, given the difference in scales at which the two sub--systems evolve, the two sub--systems are typically (re)solved separately. Popular approaches such as the {\em Flamelet Generated Manifolds} (FGM) use a two--step strategy where the governing reaction kinetics are pre--computed and mapped to a low--dimensional manifold, characterized by a few reaction progress variables (model reduction) and the manifold is then ``looked--up'' during the run--time to estimate the high--dimensional system state by the flow system. While existing works have focused on these two steps independently, in this work we show that joint learning of the progress variables and the look--up model, can yield more accurate results. We build on the base formulation and implementation \cite{ChemTab} to include the dynamically generated Themochemical State Variables (Lower Dimensional Dynamic Source Terms). We discuss the challenges in the implementation of this deep neural network architecture and experimentally demonstrate it's superior performance.
\end{abstract}


\section{Introduction}\label{introduction}
Modeling of turbulent flow combustion is central in the development of new combustion technologies in aviation, automotive and power generation~\cite{Giusti:2019}. Turbulent flow combustion combines two nonlinear and multi--scale phenomena: {\em turbulent flow} and {\em chemical reactions}. This coupling of the kinetic chemical reaction equations with the set of Navier–Stokes flow equations results in a problem that is too complex to be solved, at full resolution, by the current computational means. Even for a simple fuel such as methane, the combustion chemistry mechanism involves 53 species and 325 chemical reactions~\cite{grimech}, and the numbers increase with increasing fuel complexity. Solving the details of such mechanisms during the flow simulation can consume up to 75\% of the solution time ~\cite{ElAsrag2013ACB}. Hydrocarbon combustion, for example, involves from 50 to 7000 species depending on the fuel \cite{montgomery2005,lu2009}. Even with the aid of exascale computing, high-fidelity simulations of turbulent reactive flows with detailed kinetic remain computationally prohibitive \cite{chen2011,nouri2019}. 
\begin{figure}[H]
  \centering
  \includegraphics[width=\linewidth]{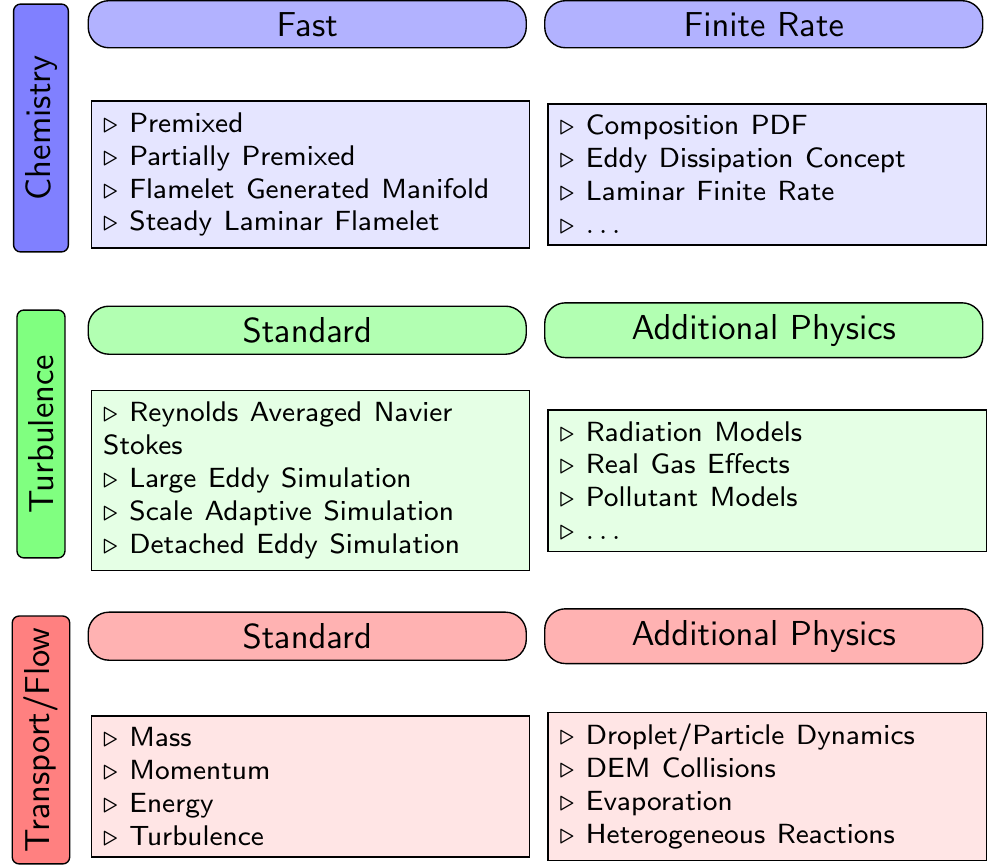}
\caption{Overview of techniques used for handling the computational complexity in coupled turbulent flow combustion systems. }\label{fig:combustionchallenges}
\end{figure}

In most cases, the large scale separation between the combustion chemistry/flame (typically sub millimeter /microsecond scale) and the characteristic turbulent flow (typically centimeter or meter/minute or hour scale) allows simplifying assumptions to be made that enable increased computational efficiency by (re)solving chemistry and flow separately ~\cite{peters2001}. Major research has focused on decoupling the systems by the development of domain specific approximation methodologies. In figure  ~\ref{fig:combustionchallenges} we have described some major methodologies in the different domains to enable simulations of turbulent flow combustion. 

\begin{figure}[H]
 \includegraphics[width=\linewidth]{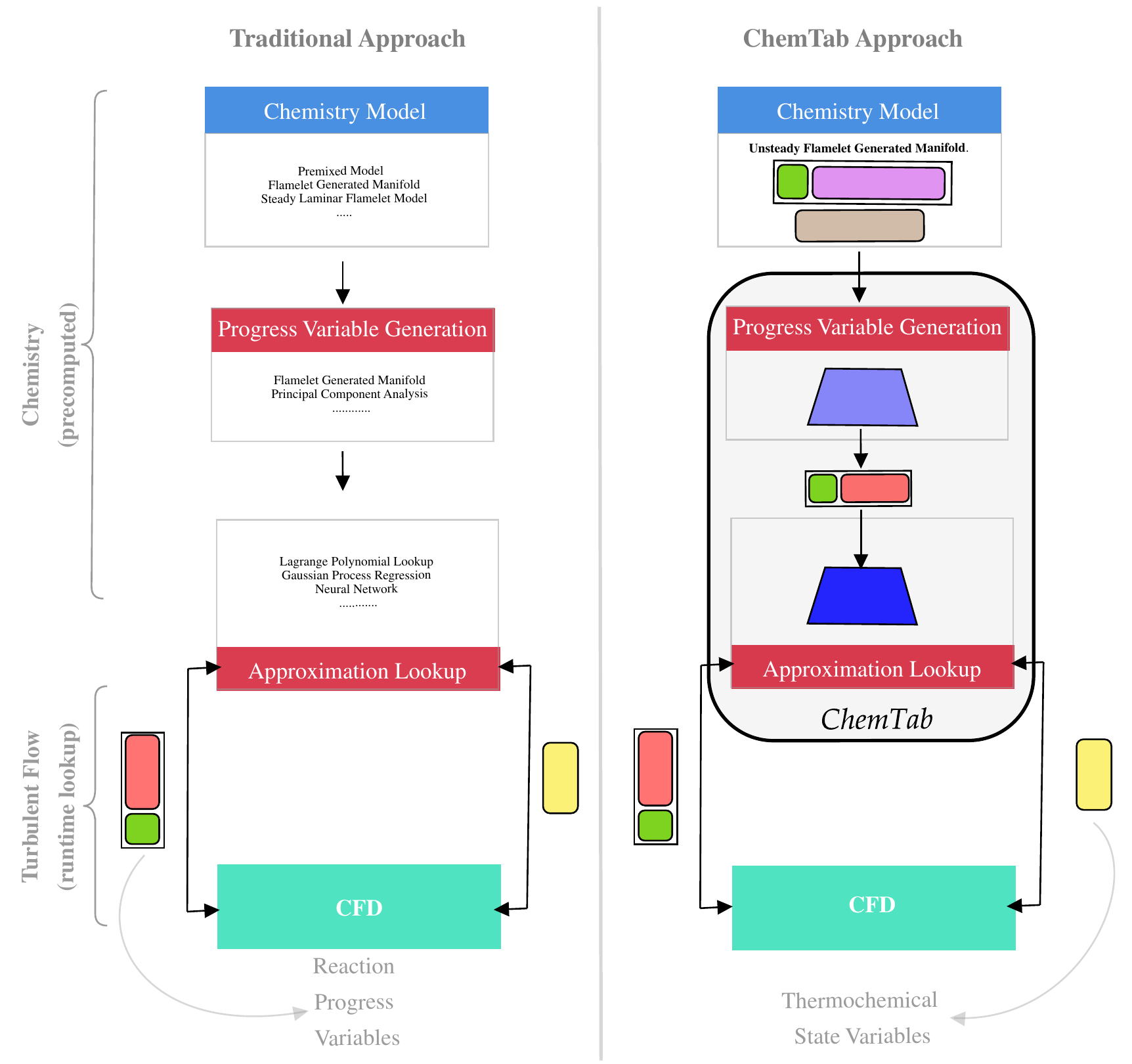}
  \caption{Traditional vs ChemTab enabled approach in a Turbulent Combustion Flow System} \label{fig:reduced}
\end{figure}

These domain specific approximations have enabled modular constructions of simulation systems. The Chemistry system is resolved first using a domain model. The solutions to the high dimensional reactions are parameterized and then stored. During the flow simulation, these solutions are looked--up to estimate the  thermochemical state, as shown in Figure~\ref{fig:reduced}.

Most of these domain specific approximations developed for increased computational efficiency rely on the existence of a theoretical low--dimensional thermochemical state--space manifold to which the combustion chemistry can be mapped~\cite{maas1992}. The central question then is, {\em how to efficiently model low--dimensional thermochemical manifolds that capture the relevant physics of the problem; and parametrize and approximate these manifolds which can then be accessed during turbulent flow simulations?} 

While existing approaches (collectively referred to as {\em state--space parametrization}~\cite{peters2001,piercemoin2004}) have been successful, they have primarily solved the two sub--problems -- {\em progress variable generation} to characterize the manifold, and {\em manifold approximation} to perform the lookup during run--time, independently. This can result in sub--optimal solutions because the progress variables, learnt using methods such as {\em Principal Component Analysis} (PCA)~\cite{sutherland20091563,biglari20154025}, are not necessarily optimized to perform the run--time lookup.  Similarly, while the traditional lookup approaches that use tabulation, or the recently proposed neural network based data--driven alternatives~\cite{bhalla2019}, facilitate efficient look--ups, the construction of the underlying data--structure or machine learning based model is not informed by the learning of the progress variables.

Our central hypothesis is that by simultaneously learning the progress variables and the manifold approximation (lookup model), we can achieve higher accuracy in terms of the estimation of the thermochemical state at run--time. But how does one combine the progress variable learning, an inherently linear mapping task, with a highly non--linear lookup model, while ensuring that the components influence each other during the learning phase? 

In a preliminary version of this work \cite{ChemTab} we proposed a framework called {\em ChemTab}, in which the learning of these two components is formulated as a joint optimization task. In this paper we extend that formulation to include newer constraints necessary for operationalization, dynamic construction of lower dimensional source terms based on the linear mapping and a novel Thermochemical State {\it Dynamic Source Term Regressor} that learns to predicts the highly nonlinear dynamic source terms and showcase the performance of this approach.  

\section{Related Work}
In this section we provide a brief overview of existing work that can be categorized into domain models, numerical/data-driven methods and deep neural networks.

\subsection{Domain Methods}
Common approaches to low--dimensional thermochemical manifold modeling are combustion chemistry mechanism reduction and thermochemical state--space parametrization ~\cite{Rastigejev13875} ~\cite{sutherland20091563}. Chemistry mechanism reduction approach cannot be generalized and in the recent past state--space parametrization approach has been the most dominant method comprising of two phases {\em progress variable generation} and {\em manifold approximation}. For progress variable generation, existing methods have either used domain models or numerical methods. 

Domain models like steady Laminar Flamelet Method (SLFM) ~\cite{PETERS1984319}, Flamelet--Generated Manifold (FGM) ~\cite{fgm2000} ~\cite{van2001}, Flamelet/Progress Variable approach (FPVA) ~\cite{piercemoin2004} ~\cite{ihme2005}  and Flamelet--Prolongation of ILDM model (FPI) ~\cite{gicquel2004} theorize that a multi--dimensional flame can be considered as an ensemble of multiple one--dimensional locally laminar flames (flamelets). These flamelets are patametrized by a combination of conserved and reactive scalars ~\cite{fgm2000} ~\cite{van2001} ~\cite{piercemoin2004} ~\cite{bojko2016}. A lot of research in this area builds on the principles laid out in ~\cite{ihme20127715} for progress variables regularization however the fundamental problem of generating adequate number of progress variables that capture the underlying physics is still open.

\subsection{Numerical Methods}
Numerical methods, like PCA, have shown significant promise for parametrization of the thermochemical state. PCA provides a method of generating reaction progress variables using the flamelet solutions, the state--space variables are still nonlinear functions of the reaction progress variables, and a nonlinear regression is learned to approximate the state--space manifold ~\cite{sutherland20091563} ~\cite{biglari20154025}~\cite{sutherland20091563} ~\cite{malik201830}~\cite{malik2020}. This purely numerical parametrization lack interpretability and may also not be generalizable enough due to variation capture maximization that may overlearn the numerical errors in the data. Linear Autoencoders have also been suggested ~\cite{Yellapantula2021} however this definition lacks incorporation of a principled approach to progress variable generation and thus may not be generalizable.

\subsection{Deep Neural Networks}
Deep Neural Networks have already achieved tremendous success in a number of domains such as computer vision and natural language processing, where large amounts of training data and highly expressive neural network architectures together give birth to solutions outperforming previously dominating methods. As a consequence, researchers have also started exploring the possibility of applying machine learning models to advance scientific discovery and to further improve traditional combustion modeling.

\subsubsection{Neural Networks for Lookup}
While domain based model have traditionally relied on tabular lookup, these are not scalable. tabulated data occupies a larger portion of the available memory on every node where the flow simulation is computing. Also the searching and retrieval of this pre--tabulated data becomes increasingly expensive in a higher--dimensional space. For example, assuming a standard 3 progress variable discretization (200, 100, 50) with say 15 tabulated thermochemical state variables, we obtain a pre--computed combustion table of 120Mb. The addition of a variable such as {\em enthalpy} with a very coarse discretization of 20 points, brings the size of the table to 2.4 Gb. To address the tabulation problem researchers like~\cite{bhalla2019} ~\cite{zhang20201} build on the work of \cite{ihme2009} to investigate the use of a neural networks for manifold approximation which replaces the Tabulation. The mapping between the progress variables (reduced dimensionality) and thermochemical state variables obtained using the flamelets solutions is learnt using a neural network. However, due to the highly non--linear, knotted and discontinuous nature of the lower dimensional manifolds formed by the progress variables generated \textit{a priori} the accuracy gained by a neural network is not satisfactory. 

\subsubsection{Physics-Informed Neural Networks (PINNs)}
Physics-Informed Neural Networks (PINNs) approximate solutions to Partial Differential Equations (PDE) by training a neural network to minimize a loss function; it includes terms reflecting the initial and boundary conditions along the space-time domain’s boundary and the PDE residual at selected points in the domain (called collocation point). PINNs are deep-learning networks that, given an input point in the integration domain, produce an estimated solution in that point of a differential equation after training \cite{raisi2018,raissi2020}. This is a promising area and the methods discussed cannot be adopted directly to solve the modeling of combustion thermo-chemistry and the parameterization of combustion manifold.

\subsubsection{Physics Guided Machine Learning}
Machine learning involve three key parts: data, model, and optimization, each of which can be integrated with prior physics knowledge. There's existing work describing physics--driven machine learning models for solving other physics problems ~\cite{willard2020,karpatne2017}, however, these methods generally focus on simpler physics and are not necessarily applicable in the domain of turbulent combustion. Our formulation is the most similar in spirit to this methodology and can be categorized as Physics-Informed Optimization.

\section{Background: Unsteady FGM}
In this work, we use the unsteady Flamelet Generated Manifolds (FGM) as the chemistry domain approximation method. FGM is a widely used tabulated chemistry method and can deal with a range of complicated conditions. FGM model shares the same theoretical basis with flamelet approaches~\cite{PETERS1984319}, in which a multi--dimensional flame can be considered as an ensemble of multiple one--dimensional flames. Generally FGM procedure used for combustion modeling follows the steps as shown below:
\begin{enumerate}
    \item Calculation of the representative 1--D flamelets.
    \item Projection of 1--D flamelets solutions Species Mass Fractions to progress variables space.
    \item Calculating the Lower Dimensional Source Terms using the projection generated above
    \item Mapping these Lower Dimensional Source Terms and other Thermo-Chemical variables to progress variables space.
    \item Generation of FGM tables according to FGM progress variables.
    \item Retrieval of thermo-chemical variables from the FGM tables according to FGM progress variables from Computational Fluid Dynamics (CFD) simulations.
\end{enumerate}

\paragraph{Notation}
The definition of terms used in the subsequent sections is provided in Table~\ref{tab:maesource}. We use bold upper-case letters to denote vectors (e.g., {\bf Y}) and use subscripts (e.g., $Y_i$) to denote the $i^{th}$ entry of the corresponding vector. Matrices are denoted using calligraphic letters (e.g., $\mathcal{Y}$).  
\subsection{Governing Equations}\label{subsec:1Dflamesol}
\begin{table}[H]
  \caption{Definitions for terms used in Section~\ref{subsec:1Dflamesol}. Terms are scalars unless noted otherwise.} 
  \begin{tabular}{ll}
   \makecell {
        \begin{tabular}{|l|l|}
            \hline
            Symbol & Description\\
            \hline
            $Z_{mix}$ & Mixture Fraction\\
            ${\bf C_{pv}}$ & Progress Variable (vector)\\
            ${\bf Y}$ & Species Mass Fraction (vector)\\    
            $\dot{\bf S}$ & Species Source Terms (vector)\\
            ${\bf h^0_{f}}$ & Heat of Formation for Species (vector)\\
            $T$ & Mixture Temperature\\
            ${\bf D}$ & Diffusivity for Species (vector)\\
            $x$ & Position in the 1--D coordinate\\
            $t$ & Time\\
            $u_x$ & Velocity along the $x$ dimension\\
            \hline
        \end{tabular}
    }
    \makecell {
        \begin{tabular}{|l|l|l|l|l|}
            \hline
             Symbol & Description\\
            \hline
            $\kappa$ & Thermal Conductivity\\
         	$Pr$ & Prandtl number\\
            $Sc$ & Schmidt number\\
            $Le$ & Lewis number \\
            $\mu$ & Viscosities \\
            $h$ & Total Enthalpy\\
            $s$ & \# Species in Mechanism\\
            $p$ & \# Progress Variables\\
                        $\rho$ & Mixture Density\\
           \hline
        \end{tabular}
    }   
  \end{tabular}
  \label{tab:maesource}
\end{table}
Conservation equations for mass, species, momentum and energy for the 1--D, fully compressible, and viscous flames, are given by:
\begin{eqnarray}
\frac{\partial \rho}{\partial t} + \frac{\partial \left(\rho u_x \right)}{\partial x} & = & 0 \label{eqn:1DflameA}\\
    \frac{\partial \left( \rho Y_i \right)}{\partial t} + \frac{\partial \rho u_x Y_i}{\partial x} & = & \frac{\partial}{\partial x}\left( \rho D_i \frac{\partial Y_i}{\partial x} \right) + \dot{S_i} \\\label{eqn:1Dflameb}
    \frac{\partial \left (\rho u_x \right)}{\partial t} + \frac{\partial \left( \rho u_x^2 \right)}{\partial x} & = & -\frac{\partial p}{\partial x} + \frac{\partial}{\partial x} \left(\mu \frac{\partial u_x}{\partial x} \right) \\\label{eqn:1Dflamec}
    \frac{\partial \left(\rho e_t \right)}{\partial t} + \frac{\partial}{\partial x} \left(\rho u_x H_t \right) & = & \frac{\partial}{\partial x} \left(u_x \mu \frac{\partial u_x}{\partial x} \right) + \mu \frac{c_p}{Pr} \left( 1 - \frac{1}{Le} \right)\frac{dT}{dx} \label{eqn:1DflameD}\\
            & + &\frac{1}{Sc} \frac{dh}{dx} - \sum \dot{S_i} h^o_{f,i}\nonumber
\end{eqnarray}
where the different terms are defined in Table~\ref{tab:maesource}.

We simplify the above equations making some well known assumptions. In 1D cartesian coordinates, the steady state solution to~\eqref{eqn:1DflameA}--\eqref{eqn:1DflameD} is obtained only when the total mass flux is zero, i.e., velocity field is zero $(u_x = 0)$ and so the four equations reduce to:
\begin{eqnarray}\label{eqn:energysteadymain}
        \frac{\partial}{\partial x} \left( \rho D_{i} \frac{\partial Y_i}{\partial x} \right) + \dot{S_i} & = & 0 \\\label{eqn:speciessteady}
        \frac{\partial}{\partial x} \left(\kappa \frac{\partial T}{\partial x} + \sum \rho D_{i} \frac{\partial Y_i}{\partial x} h_i \right) - \sum \dot{S_i} h^o_{f,i} & = & 0\label{eqn:energysteady}
\end{eqnarray}

In~\eqref{eqn:energysteady}, the final term in the energy equation is represented by the total sum of the product of all the source species and their respective heat of formation and is collectively called the source energy and will be denoted as $S_e$. Source energy is one of the crucial parameters in the combustion simulation and accurate chemistry description is required to define it. Prediction error of this term is used as the basis of comparison of our method against the other state of the art methods.

\subsection{Flamelet Solutions}
We assume that we have an ensemble of flamelets data generated by solving 1--D Steady State Flamelets differential equations in~\ref{eqn:energysteady} using a finite volume PDE solver. For each flamelet, we have access to the species mass fractions (${\bf Y}$), the thermochemical state variables ($\dot{\bf S}$ and $S_e$), and the corresponding mixture fraction ($Z_{mix}$), which are generated using the solver. We denote the collection of flamelets as matrices $\mathcal{Y}$ and $\dot{\mathcal{S}}$, and vectors $S_e$ and ${\bf Z_{mix}}$:
{
\begin{equation}
        \mathcal{Y} = \begin{bmatrix}
            Y_{11} &..&..&Y_{1s}\\
            ..&..&..&..\\
            ..&..&..&..\\
            ..&..&..&..\\
            Y_{n1}&..&.. & Y_{ns}
        \end{bmatrix},\quad
        \dot{\mathcal{S}} = \begin{bmatrix} 
            S_{11} &..&..&S_{1s}\\
            ..&..&..&..\\
            ..&..&..&..\\
            ..&..&..&..\\
            S_{n1}&..&.. & S_{ns}
        \end{bmatrix},\quad
        {\bf S_e} = \begin{bmatrix}
            S_{e_{1}}\\
            ..\\
            ..\\
            ..\\
            S_{e_{n}}
        \end{bmatrix},\quad
        {\bf Z_{mix}} = \begin{bmatrix}
            Z_{mix_{1}}\\
            ..\\
            ..\\
            ..\\
            Z_{mix_{n}}
        \end{bmatrix}
\label{eqn:datagenerated}
\end{equation}
}
Each row above corresponds to data from one flamelet (Flamekey) at a particular $x$ position (Xpos). 

\section{ChemTab Formulation \& Implementation}\label{sec:extended-formulation-implementation}
In this section we discuss how the data generated by the flamelet solver as described in \ref{eqn:datagenerated} can be used to jointly learn the progress variables and the manifold approximation by ChemTab. 

In the ChemTab aided approach, the unsteady FGM approach is replaced with the following three steps:
\begin{enumerate}
    \item Calculation of the representative 1D flamelets (data generation)
    \item Using the data generated jointly generate Progress Variables  (encoder) and Manifold Approximation (regressor(s)) using ChemTab
    \item Retrieval of thermo--chemical variables from the ChemTab--regressor(s) according to progress variables from CFD simulations.
\end{enumerate}
As can be noted several of the dis-jointed steps of the unsteady FGM procedure are replaced by our unified formulation which enables greater accuracy and ease of deployment as a key module of the overall turbulent combustion flow system. 

\subsection{Formulation}\label{sec:extended-formulation}
This section presents the formulation that focuses on creating a linear encoder for progress variable generation that is influenced by source energy and some of the key higher dimensional source Terms. Source energy is a key thermochemical state variable along with lower dimensional Source Terms that are needed at run--time by the CFD solver so we include those as a key output(s) of the regressor(s) as a part of the formulation. 

The functional relationship between ${\bf Y},{\bf \dot{S}}, S_{e}$ is described by \ref{eqn:energysteadymain}, from a machine learning standpoint we can learn these functional approximations from the data \ref{eqn:datagenerated}. These functional approximations conceptualize the relationships between Species Mass Fractions, Source Terms and Source Energy as follows:
\begin{align}
    \dot{\bf S} & = \Phi({\bf Y})\\
    S_{e} & = - \sum_{i}^{s} h_{f,i}^{0} * \dot{S_{i}} = \Psi({\bf Y})
    \label{eqn:conceptual--energysteady}
\end{align}  

When we linearly embed ${\bf Y}$ into the lower dimensional ${\bf C_{pv}}$ the functional relationship will change and can be conceptualized as follows: 

\begin{align}
    \dot{\bf S} & = \phi({\bf C_{pv}})\\
    S_{e} & = \psi({\bf C_{pv}})
    \label{eqn:conceptual-II-energysteady}
\end{align}

We now present our formulation which jointly addresses the three learning problems. {\em Encoder} ($\omega$) that projects the higher dimensional Species Mass Fractions ${\bf Y}$ to a lower dimensional manifold to create the lower dimensional ${\bf C_{pv}}$--the linear embedding/progress variables. The {\em {\it Physics Regressor}} which learns the relationships ($\phi$, $\psi$) between the progress variables $({\bf C_{pv}}, Z_{mix})$ and key Source Terms $\dot{\bf{S}}$ and the most import thermochemical state variable $S_e$ respectively. And finally the {\em {\it Dynamic Source Term Regressor}} which learns the relationship $\gamma$ between the progress variables $({\bf C_{pv}}, Z_{mix})$ and lower dimensional Source Terms $\tilde{\dot{\bf{S}}}$. The lower dimensional Source Terms are constructed using the same Encoder used for the species mass fractions. 

Clearly all of these learning problems are interrelated. The {\it Physics Regressor} by the virtue of learning $\phi$, $\psi$ influences $\omega$. The {\it Dynamic Source Term Regressor} also by the virtue of learning $\gamma$ influences $\omega$. To account this  inter-relatedness we use a multi-objective optimization formulation as follows.

\begin{subequations}\label{eqn:chemtabdynsrcformulation}
    \begin{alignat}{1}
        \min \quad \alpha_{1}\sum_{j=1}^{n} \mathcal{L}_{phy_1}(S_{e},\psi({\bf C_{pv}},Z_{mix})) \quad+ \alpha_{2}\sum_{i=1}^{s}\sum_{j=1}^{n} \mathcal{L}_{phy_2}({\bf \dot{S}_{ij}},& \phi({\bf C_{pv}},Z_{mix})) \quad+ \alpha_{3}\sum_{i=1}^{p}\sum_{j=1}^{n} \mathcal{L}_{dsreg}({\bf \tilde{\dot{S_{ij}}}},\gamma({\bf C_{pv}},Z_{mix}))\\ 
        &\textrm{s.t.}\\ 
            &\def\sss{\scriptscriptstyle}
            \setstackgap{L}{8pt}
            \def\stacktype{L}
            \stackunder{\mathrm{{\bf C_{pv}}}}{\sss p} = \omega({\bf Y}) = \stackunder{{\bf Y}}{\sss s} \times \stackunder{\mathcal{W}}{\sss s\times p} \label{eqn:linearity-constraint} \\
            &p << s\label{eqn:dimreduction}\\
            &\mathcal{W^{T}} \times \mathcal{W} = I \label{eqn:ext-WO-constraint} \\
            &\mathcal{W}_{ic} > 0 \quad \forall \quad i,c\label{eqn:ext-NN-constraint}\\
            &({\bf C_{pv}} \oplus Z{mix})^{T} \times ({\bf C_{pv}} \oplus Z{mix}) \approx I \label{eqn:ext-AR-constraint}\\
            &\def\sss{\scriptscriptstyle}
            \setstackgap{L}{8pt}
            \def\stacktype{L}
            \stackunder{\mathrm{\tilde{\dot{\bf{S}}}}}{\sss p} =\omega({\bf \dot{S}}) =  \stackunder{{\bf \dot{S}}}{\sss s} \times \stackunder{\mathcal{W}}{\sss s\times p} \label{eqn:same-linearity-constraint-dynsourceterms}
\end{alignat}
\end{subequations}    

Each of the Loss function $\mathcal{L}$ in \ref{eqn:chemtabdynsrcformulation} serves as a method of evaluating how well the learning is across the $n$ data points and $\alpha_{1}$,$\alpha_{2}$, $\alpha_{3}$ are the contributions of the individual loss functions to the overall objective. These weights can be either solved for as an hyper-parameter or can be a design choice for the subject matter expert.

$\mathcal{L}_{phy_1}$ captures the loss between the true $S_{e}$ and the predicted $\hat{S_{e}}$ across all of the $n$ data points. $\mathcal{L}_{phy_2}$ captures the loss between the true ${\bf \dot{S}}_{i}$ and the predicted ${\hat{\bf \dot{S}}}_{i}$ for all the source species across all of the $n$ data points. A slight variation would be to use $k$ key source terms instead of all the $s$ source terms. $\mathcal{L}_{dsreg}$ captures the loss between the dynamically created true lower dimensional source terms ${\tilde{\bf \dot{S}}}_{i}$ and the predicted $\hat{{\tilde{\bf \dot{S}}}}_{i}$.

The constraints \ref{eqn:linearity-constraint} and \ref{eqn:dimreduction} ensures that the embedding $\omega$ is linear dimensionality reduction. The constraint \ref{eqn:ext-WO-constraint} ensures that the linear projection matrix is Orthogonal and the constraint \ref{eqn:ext-AR-constraint} ensures that the progress variables generated from this embedding are orthogonal. These constraints together are Principal Component Analysis (PCA) inspired. And the \ref{eqn:ext-NN-constraint} constraint ensures every entry in the matrix is positive, this is a necessary constraint for the productionalization of our work. The constraint \ref{eqn:same-linearity-constraint-dynsourceterms} ensures that the dynamic source terms $\tilde{\dot{S}}_{i}$ are constructed using the same encoder. 

\subsection{Implementation}\label{sec:extended-implementation}
We present a Deep Neural Architecture implementation of the joint optimization formulation discussed in the previous section. The implementation assumes the input data is in the structure described in ~\eqref{eqn:datagenerated}. The construction of the Deep Neural Network is presented below:
    
\begin{subequations}\label{eqn:DNN-dynsrc--expansion}
    \begin{alignat}{1}    
    f_{\mathcal{\theta}}^{[0]}(y) &= y \label{eqn:dnn-first-layer}\\
    f_{\mathcal{\theta}}^{[1]}(y) &= {\bf C_{pv}} =  (W^{[0]} f_{\mathcal{\theta}}^{[0]}(y)) \label{eqn:dnn-second-layer}\\
    f_{\mathcal{\theta}}^{[2]}(y) &= (f_{\mathcal{\theta}}^{[1]}(y) \oplus Z_{mix}) \label{eqn:dnn-concat-layer}\\
    f_{\mathcal{\theta}}^{[l]}(y) &= \mathcal{\sigma} \: \mathcal{\rm o} \: (W^{[l−1]} f_{\mathcal{\theta}}^{[l-1]}(y) \:+\: b^{[l-1]} ) \:\:\: \forall \quad l \quad \textrm{s.t.} \quad {3 \leq l \leq L-1}  \label{eqn:dnn-post-concat-layer-to-last}\\
    f_{\mathcal{\theta}}(y) &= f_{\mathcal{\theta}}^{[L]}(y) = \mathcal{\sigma} \: \mathcal{\rm o} \: (W^{[L-1]} f_{\mathcal{\theta}}^{[L-1]}(y) \:+\: b^{[L-1]} )\label{eqn:dnn-phy-reg-output}\\
    \tilde{\dot{{\bf S}}} &= f_{\mathcal{\theta}}^{[1]}(\dot{S})\label{eqn:dnn-dyn-source-creation}\\
    g_{\mathcal{\theta}}^{[0]}(y) &= f_{\mathcal{\theta}}^{[2]}(y)\label{eqn:dnn-dyn-source-first-layer} \\
    g_{\mathcal{\theta}}^{[l]}(y) &= \mathcal{\sigma} \: \mathcal{\rm o} \: (W^{[l−1]} g_{\mathcal{\theta}}^{[l-1]}(y) \:+\: b^{[l-1]} ) \:\:\: \forall \quad l \label{eqn:dnn-dyn-src-layers}\\
    g_{\mathcal{\theta}}(y) &= g_{\mathcal{\theta}}^{[L]}(y) = \mathcal{\sigma} \: \mathcal{\rm o} \: (W^{[L-1]} g_{\mathcal{\theta}}^{[L-1]}(y) \:+\: b^{[L-1]} )\label{eqn:dnn-dyn-src-reg-output}
\end{alignat}
\end{subequations} 
In the dnn architecture \ref{eqn:DNN-dynsrc--expansion} the first layer \ref{eqn:dnn-first-layer} accepts the species mass fractions $y$. The second layer \ref{eqn:dnn-second-layer} is the {\em Encoder} which generates the projection matrix $\mathcal{W}$ used to create the progress variables $\tilde{Y}$.This layer has a linear activation function. The third layer \ref{eqn:dnn-concat-layer} concatenates $\tilde{Y}$ and $Z_{mix}$. The {\em Physics Regressor} is composed of the subsequent layers of the $f_{\theta}$ network. The layers in the {\em Physics Regressor}  \ref{eqn:dnn-post-concat-layer-to-last} use a nonlinear activation function. And the last layer \ref{eqn:dnn-phy-reg-output} network generates the source energy $S_{e}$ and the key source terms $\dot{S_i}$. 

The $g_{\theta}$ network is the {\it Dynamic Source Term Regressor}. The \ref{eqn:dnn-dyn-source-creation} creates the true values for the Dynamic Source Terms $\tilde{\dot{S}}$. The \ref{eqn:dnn-dyn-source-first-layer} accepts the progress variables concatenated in \ref{eqn:dnn-concat-layer}.  The {\it Dynamic Source Term Regressor} is composed of the subsequent layers of the $g_{\theta}$ network. The layers in the {\it Dynamic Source Term Regressor}  \ref{eqn:dnn-dyn-src-layers} use a nonlinear activation function. The last layer \ref{eqn:dnn-dyn-src-reg-output} generated the predicted lower dimensional dynamic source terms $\hat{\tilde{\dot{S}}}$.

The Deep Neural Network can be then trained using the following loss function and layer constraints:
\begin{equation}\label{eqn:DNN-ChemTab-dynsrc--loss}
    \begin{aligned}
        \arg\min_{\mathcal{\theta}} \quad \mathcal{L} (f_{\mathcal{\theta}}(y), \mathcal{S})  + \mathcal{L}( g_{\mathcal{\theta}}(y) , \tilde{\dot{{\bf S}}}) \\
        s.t. \quad W^{[0]T}W^{[0]} = I\\
        W_{ic} > 0  \quad \forall \quad i,c \\
        f_{\mathcal{\theta}}^{[2]}(y)^Tf_{\mathcal{\theta}}^{[2]}(y) \approx I\\
        \quad \dot{{\bf S}} \times W^{[0]} = \tilde{\dot{{\bf S}}}
    \end{aligned}
\end{equation} 

Where $\mathcal{S}= \{S_e, \dot{S}_i \} $ for $k$ important species. The $\mathcal{L}$ loss functions can take several variations, Mean Absolute Error (MAE) , Sum of Squared Error(SSE) and $-R^{2}$ being the common choices.   

\begin{equation}\label{eqn:r2-metric}
    \begin{aligned}
    R^{2} = 1-\frac{\sum_{j=0}^{n}(S-\widehat{S})^2}{\sum_{i=0}^{n}(S-\bar{S})^2}
    \end{aligned}
\end{equation}

The common $R^2$ metric is defined in the equation \ref{eqn:r2-metric}. This metric focuses on maximizing the variance capture.

\begin{equation}\label{eqn:DNN-ChemTab-r2-loss}
    \begin{aligned}
    \min \quad -1 * \left( \underbrace{\frac{1}{k+1}\left(1-\frac{\sum_{j=0}^{n}({S}_{e_{j}}-\hat{S}_{e_{j}})^2}{\sum_{j=0}^{n}({S}_{e_{j}}-\bar{S}_{e_{j}})^2} \quad + \quad  \sum_{i=0}^{k}{1-\frac{\sum_{j=0}^{n}(\dot{S}_{ij}-\hat{{\dot{S}}}_{ij})^2}{\sum_{j=0}^{n}({\dot{S}}_{ij}-\bar{{\dot{S}}}_{ij})^2}} \quad\right)}_{\textrm{{\it Physics Regressor Loss}}} + \underbrace{\quad \frac{1}{p}\sum_{i=0}^{p}{1-\frac{\sum_{j=0}^{n}(\tilde{\dot{S}}_{ij}-\hat{\tilde{\dot{S}}}_{ij})^2}{\sum_{j=0}^{n}(\tilde{\dot{S}}_{ij}-\bar{\tilde{\dot{S}}}_{ij})^2}}\quad}_{\textrm{{\it Dynamic Source Term Regressor Loss}}} \right)
    \end{aligned}
\end{equation}

In the equation \ref{eqn:DNN-ChemTab-r2-loss} we present the loss function we used for our DNN implementation. The first term of the loss function measures the loss of the {\it Physics Regressor} and the second term the loss of the {\it Dynamic Source Term Regressor}. For the {\it Physics Regressor} we choose the average $R^2$ metric across Source Energy $S_{e}$ and the $k$ key higher dimension Source Terms $\dot{S_{i}}$. For the{\it Dynamic Source Term Regressor} we choose the average $R^2$ metric across $p$ Dynamic Source Terms $\tilde{\dot{S}}_{i}$.

\begin{figure}[H]
  \centering
  \includegraphics[width=\linewidth]{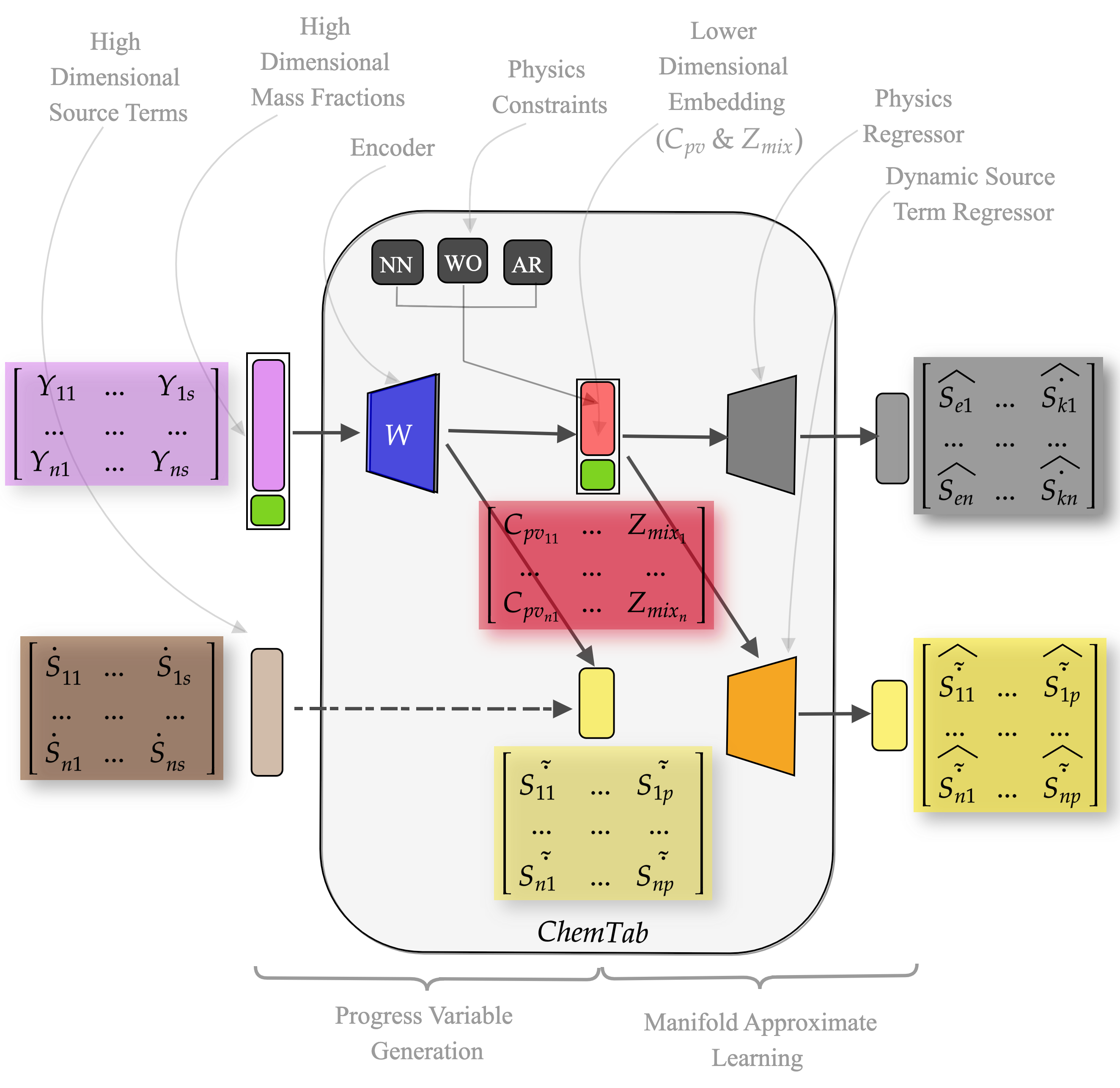}
  \caption{ChemTab Extended Architecture and Training Procedure}
  \label{fig:chemtab-extended-architecture}
\end{figure}
 
The figure \ref{fig:chemtab-extended-architecture} describes the architecture and the process used to  
train the model. Of note is the construction of true values of the dynamic source terms $\tilde{\dot{S}}_{i}$ that are used to train the {\it Dynamic Source Term Regressor}. For each batch iteration we create these values based on the current value of the projection matrix $\mathcal{W}$. Essentially these values will be lagging one iteration behind the {\it Physics Regressor}. As the training stabilizes the updates in the projection matrix will plateau and the values of true values dynamic source terms will stabilize. 

\section{Experimentation \& Results}
In this section we explain the specifics of the data set creation, the data sets used for evaluation, the training strategy, some interesting results and the performance of the implementations of the various formulation in the context of the multiple objectives.

\subsection{Dataset Generation}\label{sec:datageneration}
The training data was generated by solving 1--D Steady State Flamelets differential equations using a finite volume PDE solver. To model the chemical kinetics reaction rates, a variety of mechanisms are adopted in the combustion community. Depending on the hydrocarbon fuel different mechanisms are chosen which closely describe the chemistry associated with the fuel of simulation. Methane is the basic hydrocarbon and one of the major products of many higher order hydrocarbons. GRI--Mech 3.0 is one of the widely used Methane mechanism to model the reaction kinetics. This mechanism consists of 53 chemical species and 325 reactions. 

The Flamelet solver discretizes the domain into $200$ grid points (200 observations on the axial coordinate) in between the fuel and the air boundary and $100$ flame are solved to steady--state. Once the solution reaches steady--state the solver completes one iteration. For the next iteration flame solution is strained by reducing the domain by $0.99$ and the process is continued until the flame extinguishes. Each flame is then tagged with the corresponding strain rate that is called a flame--key. To train the model 20,000 data points (100 flames and 200 grid points) for a single pressure setting are used. The data is generated using an in--house solver which creates the flame solutions and stores the required data. Some of the generated data represents extinguished flames, we choose to retain this data in our model training, this becomes an extremely challenging task as there are flames that show large amounts of activity and the model has to learn this phenomenon.

\subsection{Implementation and Settings}
 
We implemented ChemTab using Tensorflow 2.3.0, Keras and Adam optimizer. Models were trained on a server with Nvidia Quadro RTX 5000 GPU and cuDNN 8.0 and CUDA 11.0.

\subsubsection{Hyper-Parameters Investigated}
\begin{table}[H]
 {\small
  \caption{Hyper-Parameters}
  \label{tab:ChemTabhyper-parameters}
  \begin{tabular}{|ll||ll|}
    \hline
    Parameter & Range & Parameter & Range\\
    \hline
 	Largest Layer Width & 128 - 4096 & Number of $C_{pv}$ & 4 - 12\\
    Dropout  & 0\% - 40\% & Activation Functions & ReLU, TanH, SeLU\\
    Batch Size & 128 - 1028 & Output Scaler & MinMaxScaler, RosbustScaler\\ 
   \hline
\end{tabular}\label{tbl:hyper-parameters}}
\end{table}

We performed bayesian optimization on the several hyper-parameters mentioned in the \ref{tbl:hyper-parameters}. The hyper-parameter optimization module sampled several values for the range of the hyper-parameter.

\begin{figure}[H]
  \centering
  \includegraphics[width=0.6\linewidth]{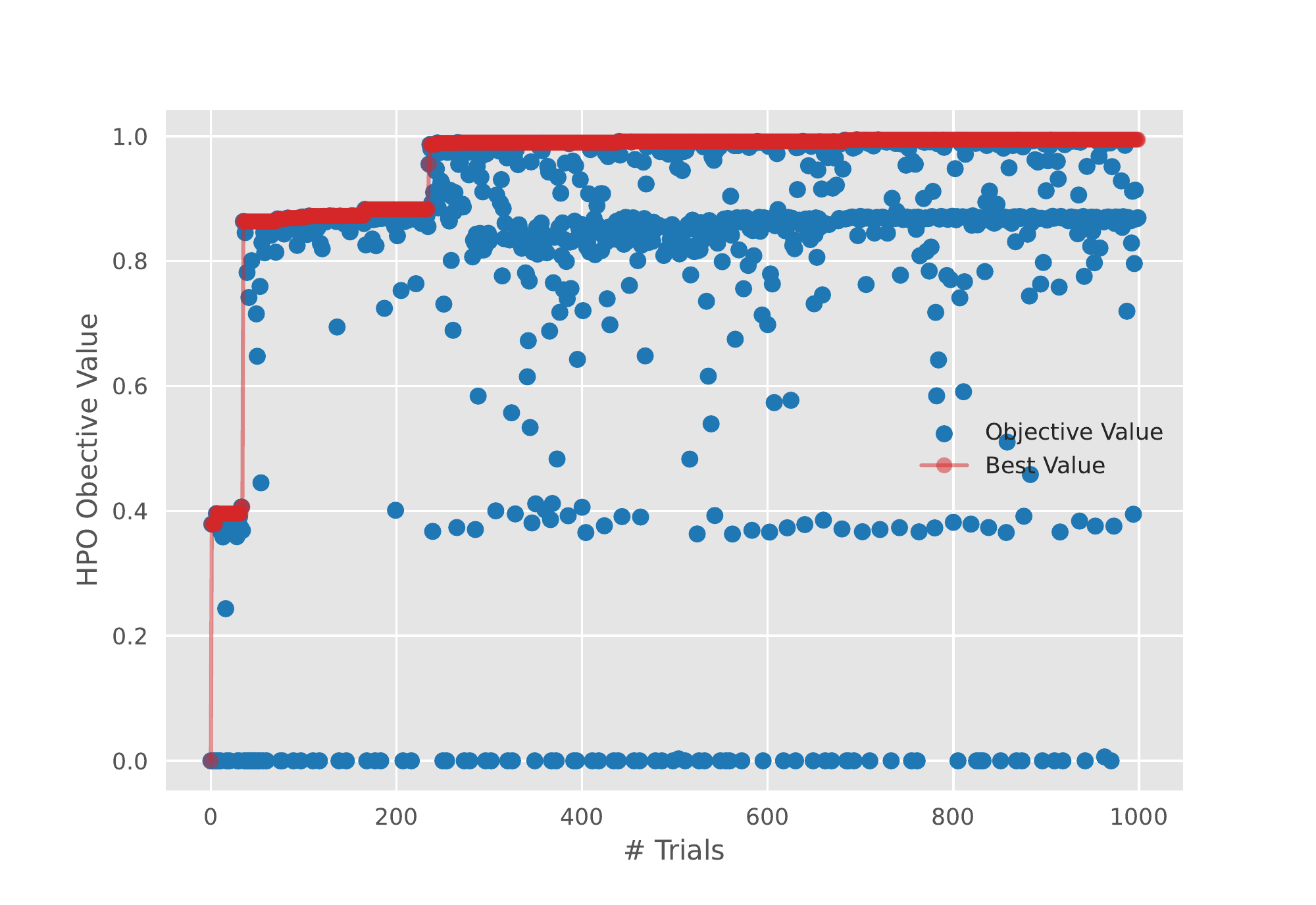}
  \caption{}
  \label{fig:no-of-trials-hpo}
\end{figure}

As can be seen in \ref{fig:no-of-trials-hpo} it took the hyper-parameter optimizer several hundred trials before it started reaching diminishing returns. This indicates the hyper-parameter settings we are changing are non-trivial. A 'trial' in this case are the number of hyper-parameter configuration experiments it has performed at a given point.

\subsubsection{\# of Reaction Progress Variables}

Two key hyper-parameters are the number of progress variables $C_{pv}$ and the Width of the Middle Layer for the Regressor. We construct both of the regressors with a identical number of layers and the width of each layer. The Middle Layer is the largest and 4 layers on either side of it which keep decreasing by a factor of 2.  As an example if the Middle Layer width is 800 then the regressors end up looking like: {\it input,50,100,200,400,800,400,200,100,50,output}.

\begin{figure}[H]
  \centering
  \includegraphics[width=0.6\linewidth]{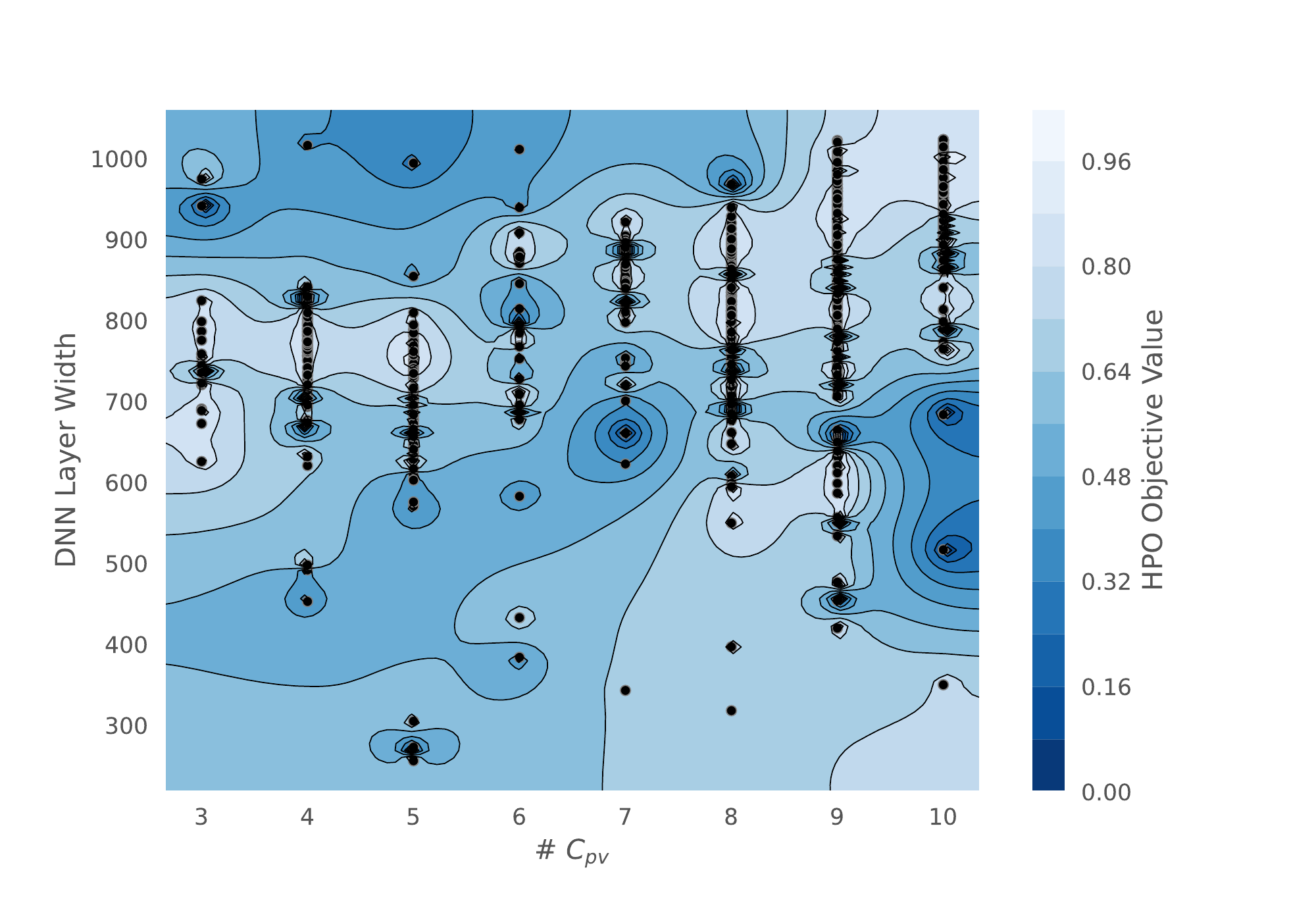}
  \caption{DNN Layer Width and no. of $C_{pv}$ HPO Objective Contour Plot}
  \label{fig:hpo-contour-layerwidth-numcpv}
\end{figure}

In the figure \ref{fig:hpo-contour-layerwidth-numcpv} we present the influence of model size and the number of $C_{pv}$ used on the model's performance. As it turns out the number of  $C_{pv}$ used didn't have a very large impact on performance (more  $C_{pv}$ = more  $C_{pv}$ source terms to predict accurately). Nonetheless we can see the highest performing regions are with high  $C_{pv}$ count is 9 and large network width 800.

\subsubsection{Final Model Configurations}
After the hyper-parameter optimization we settled on the final configurations.

\begin{table}[H]
 {\small
  \caption{Model Parameters}
  \label{tab:ChemTabParameters}
  \begin{tabular}{|ll||ll|}
    \hline
    Parameter & Value & Parameter & Value\\
    \hline
 	Learning Rate & 0.001 & Number of Layers & 11\\
 	Output-Scaler & RobustScaler & Regresssor Layer Shapes & input|49|99|198|396|792|396|198|99|49|output \\
    Dropout  & 1.522\% & Activation Functions & ReLU\\
    Early Stopping  & Yes & Number of epochs & 500\\
    Batch Size  & 407 & Network Weight Initialization & Glorot Uniform Distribution\\ 
   \hline
\end{tabular}\label{tbl:results-hyper-parameters}}
\end{table}

In the table \ref{tbl:results-hyper-parameters} we present the final values of the parameters learn't from hyper-parameter optimization and key model settings.

\subsection{Evaluation}
As discussed in \ref{sec:extended-formulation-implementation} the formulation can use several objective functions for the individual regressor(s) during the model training. We trained several variations using the $R^2$ metric based loss function as defined in \ref{eqn:DNN-ChemTab-r2-loss}. We will evaluate the overall model performance based on the same and the individual terms using $R^2$ metric as defined in \ref{eqn:r2-metric}.

\subsection{Results}

\begin{figure}[H]
  \centering
  \includegraphics[width=0.6\linewidth]{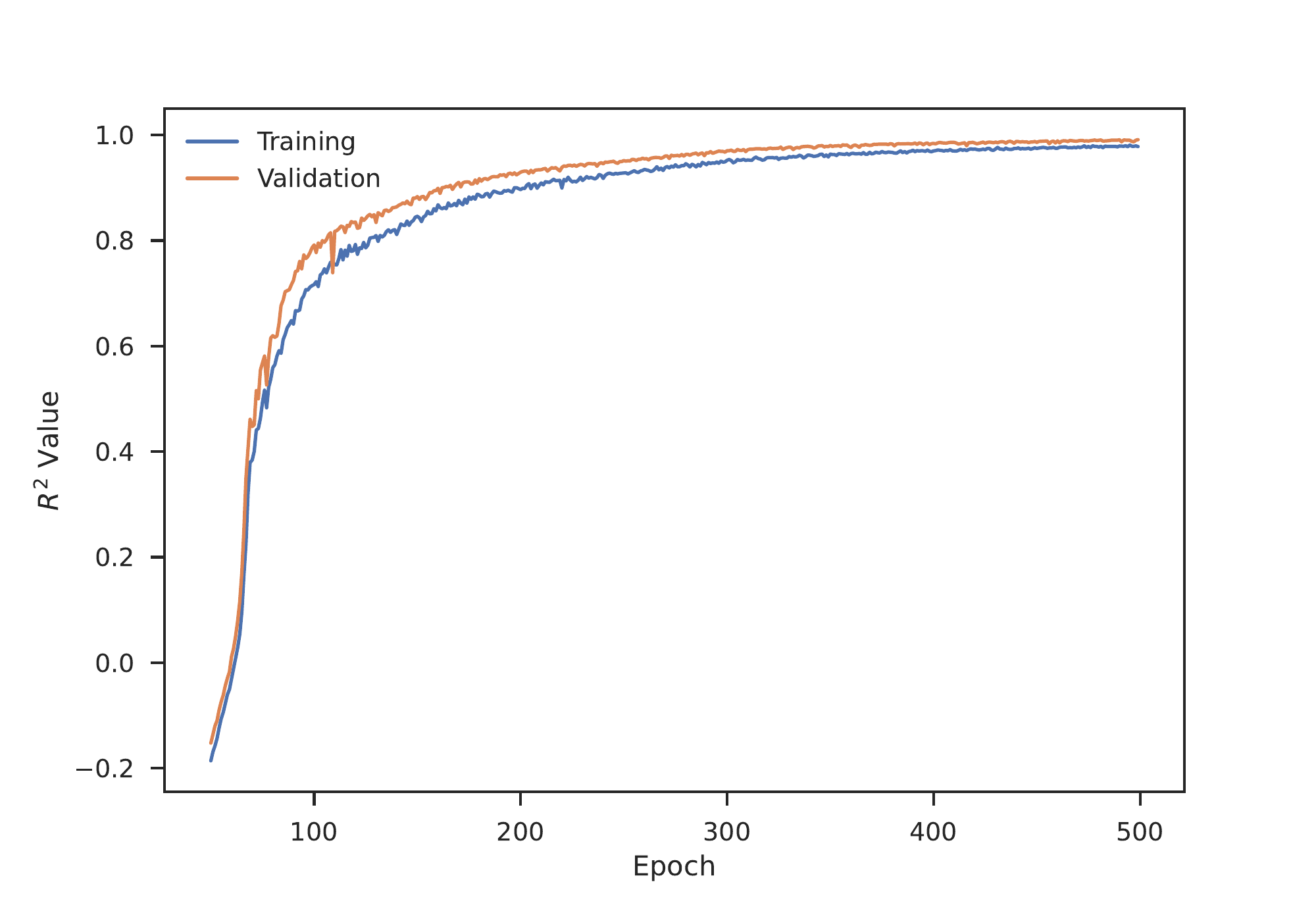}
  \caption{Model Training and Validation Performance}
  \label{fig:extended-model-loss}
\end{figure}

In the \ref{fig:extended-model-loss} we present the Model Loss function value across the training epochs (in this case loss is $-R^2$, we've made it positive for display purposes). The model performance rapidly increases across the first 125 training epochs and then is steady from epochs 125 to 300. After epoch number 300 the model improves only marginally. 

\subsubsection{Model Performance}
Here we compare the performance of the best model with the other DNN based benchmarks and the original framework benchmark.

{\it{\bf Current Framework Comparison}}
The current framework uses FGM based progress variables and Conformal Mapping based Tabulation and Lagrange Polynomial Interpolation based lookup. The tabulation was generated by using the entire data--set. The best $R^2$ for $S_{e}$ from that the framework generated on the data--set was 0.852417. The best ChemTab model trained on 50\% of the data showed a 16\% increase in $R^2$. This increase although high comes from the limitation of the current framework to include more than 2 progress variables and the realization of that through conformal mapping. We present a more principled comparison with the state--of--the--art methods in the next section.

{\it{\bf Other Baseline Comparisons}}
We now present the performance comparison of our model with the appropriate baselines. We note that we compare the performance of the {\it Physics Regressor} against the following baselines that are purely trained for just that task.

\begin{table}
  \caption{Current state of the art methods and ChemTab} \label{tab:baselines}
  \begin{tabular}{|l|c|l|}
    \hline
    Method Abbreviation & Progress Variable Generation & Manifold Approximation ($S_{e}$)\\
   \hline
    FGM--CPVG--DNN\label{method:FGM--CPVG--DNN} & \makecell[l] {FGM Constrained }  & \makecell {DNN}\\
    DNN--PVG(NL)--DNN\label{method:DNN--PVG(NL)--GP} & \makecell[l] {Non--Linear Encoder}  & \makecell {DNN}\\
    DNN--PVG(UL)--DNN\label{method:DNN--PVG(UL)--DNN} & \makecell[l] {Unconstrained Linear Encoder }  & \makecell {DNN}\\
    ChemTab & \makecell[l] {Physics constrained Linear Encoder \ref{eqn:dnn-second-layer}}  & \makecell {DNN}\\
\hline
\end{tabular}
\end{table}

\begin{table}[H]
    \caption{Source Energy and Key Species Benchmark ($R^2$) }
         \begin{tabular}{|l|l|l|l|l|}
        \hline
         Dependent & ChemTab  & DNN--PVG(UL)--DNN & DNN--PVG(NL)--DNN & FGM--CPVG--DNN \\
         \hline  
         $S_{e}$ & 0.995184 & {\bf 0.996250} & 0.992878 & {\bf 0.966763} \\
         $\dot{S}_{O2}$ & 0.996258 & {\bf 0.996672} & 0.994565 & 0.988661\\
         $\dot{S}_{CO}$ & 0.998296 & 0.996609 & 0.996406 & 0.969174 \\
         $\dot{S}_{CO2}$ & 0.996334 & {\bf 0.998193} & {\bf 0.997022} & 0.995969\\
         $\dot{S}_{H2O}$ & 0.998542 & 0.998026 & 0.993117 & 0.974178\\
         $\dot{S}_{OH}$ & 0.994757 & 0.993910 & 0.985228 & 0.969222\\
         $\dot{S}_{H2}$ & 0.997004 & 0.995364 & 0.993076 & 0.946442\\
         $\dot{S}_{CH4}$ & 0.991398 & 0.99867 & {\bf 0.997792} & 0.925120\\
        \hline
        \end{tabular}
    \label{tab:results-ext-benchmark-comparison}
\end{table}       
None of the benchmark models adhere to the operationalization constraints of \ref{eqn:linearity-constraint} , \ref{eqn:same-linearity-constraint-dynsourceterms} \ref{eqn:ext-NN-constraint} and hence can not be used however we use them as reference.  As seen in the table \ref{tab:results-ext-benchmark-comparison}tThe {\it Physics Regressor} performs better in most cases and performs adquately in others.

{\it{\bf Current Framework Comparison}}
For the best model we present the $R^2$ metric for each of the regressors
\begin{table}[H]
\centering
  \caption{Best Model $R^2$ Scores} 
  \begin{tabular}{ll}
  \makecell{{\it Physics Regressor}}
  &
  \makecell{{\it Dynamic Source Term Regressor}}
\\     
  \makecell{
        \begin{tabular}{|l|l|}
        \hline
         Dependent &       $R^2$ \\
         \hline  
         $S_{e}$ & 0.995184 \\
         $\dot{S}_{O2}$ & 0.996258 \\
         $\dot{S}_{CO}$ & 0.998296 \\
         $\dot{S}_{CO2}$ & 0.996334 \\
         $\dot{S}_{H2O}$ & 0.998542 \\
         $\dot{S}_{OH}$ & 0.994757 \\
         $\dot{S}_{H2}$ & 0.997004 \\
         $\dot{S}_{CH4}$ & 0.991398 \\
        \hline
        \end{tabular}}
  &
  \makecell{
        \begin{tabular}{|l|l|}
        \hline
                    Dependent &       $R^2$ \\
        \hline
        $\tilde{\dot{S}}_{1}$ & 0.993107 \\
        $\tilde{\dot{S}}_{2}$ & 0.998045 \\
        $\tilde{\dot{S}}_{3}$ & 0.997667 \\
        $\tilde{\dot{S}}_{4}$ & 0.997427 \\
        $\tilde{\dot{S}}_{5}$ & 0.998396 \\
        $\tilde{\dot{S}}_{6}$ & 0.994888 \\
        $\tilde{\dot{S}}_{7}$ & 0.998462 \\
        $\tilde{\dot{S}}_{8}$ & 0.995614 \\
        $\tilde{\dot{S}}_{9}$ & 0.997340 \\
        \hline
        \end{tabular}}
  \end{tabular}
  \label{tab:results-r2-scores-methane}
 \end{table} 

As can be observed in the table \ref{tab:results-r2-scores-methane}, the model does an excellent job of capturing the variance across the entire dataset.

\begin{figure}[H]
  \centering
  \includegraphics[width=1.0\linewidth]{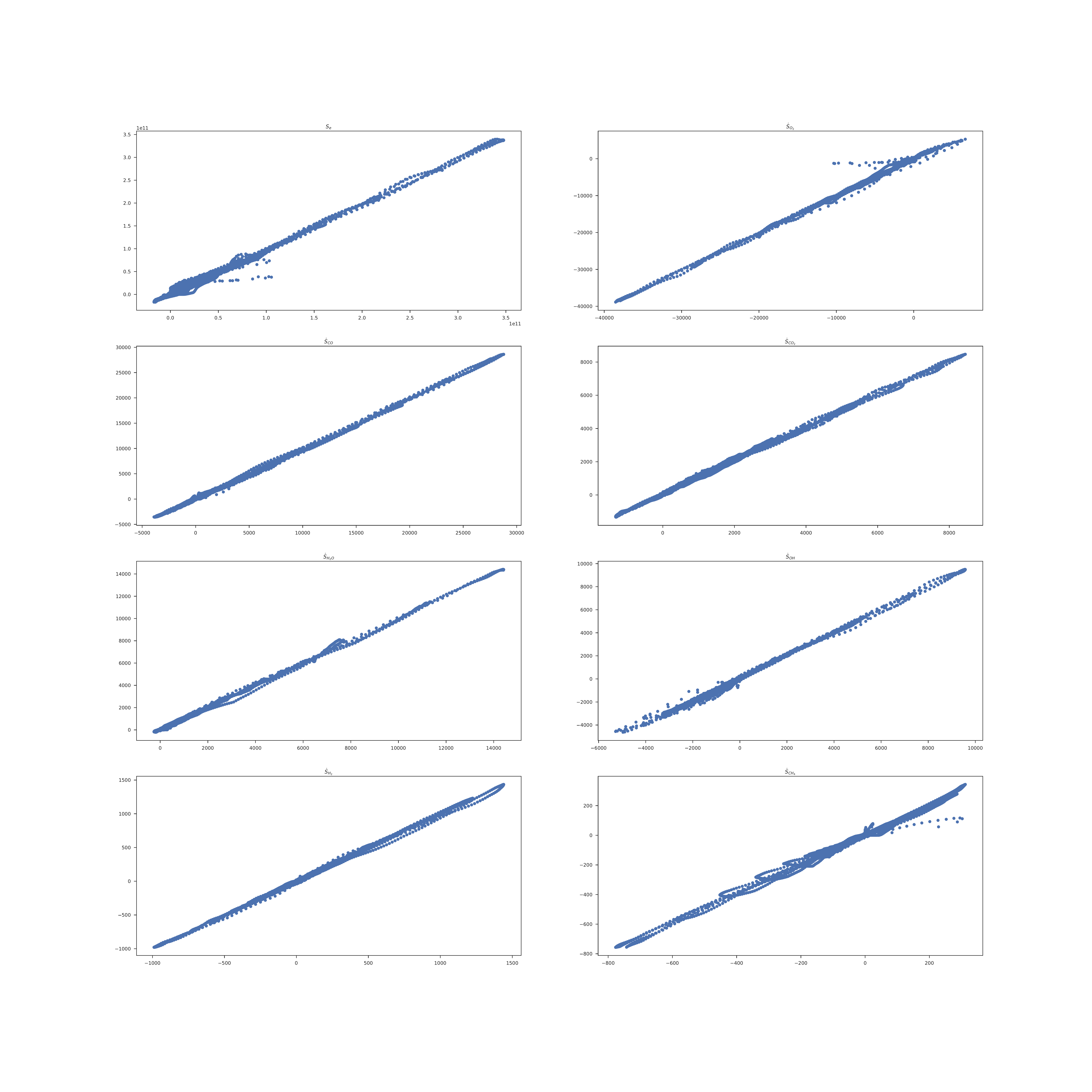}
  \caption{True Value vs Predicted Value Plots - Physics Regressor}
  \label{fig:qq-static-source-prediction}
\end{figure}

In the figure \ref{fig:qq-static-source-prediction} the true value is plotted on the X-axis and the prediction is plotted on the Y-axis. It can be observed that the {\it Physics Regressor} model does extremely well and is able to capture most of the variability across the Source Energy ($S_{e}$)/souener except for the range $0.5e^{11}$ and $1.0e^{11}$. The model under-predicts for these values. The model over-predicts for $S_{O_{2}}$ for the ranges between -1000 and 0. The model overall does a decent job for $S_{CH_{4}}$ but over or under predicts marginally through the range. This is expected from the source term for the fuel Methane($CH_{4}$) as there will be more variability compared to other source terms. This will need further investigation and a stratified oversampling strategy per batch may alleviate this issue.

\begin{figure}[H]
  \centering
  \includegraphics[width=1.0\linewidth]{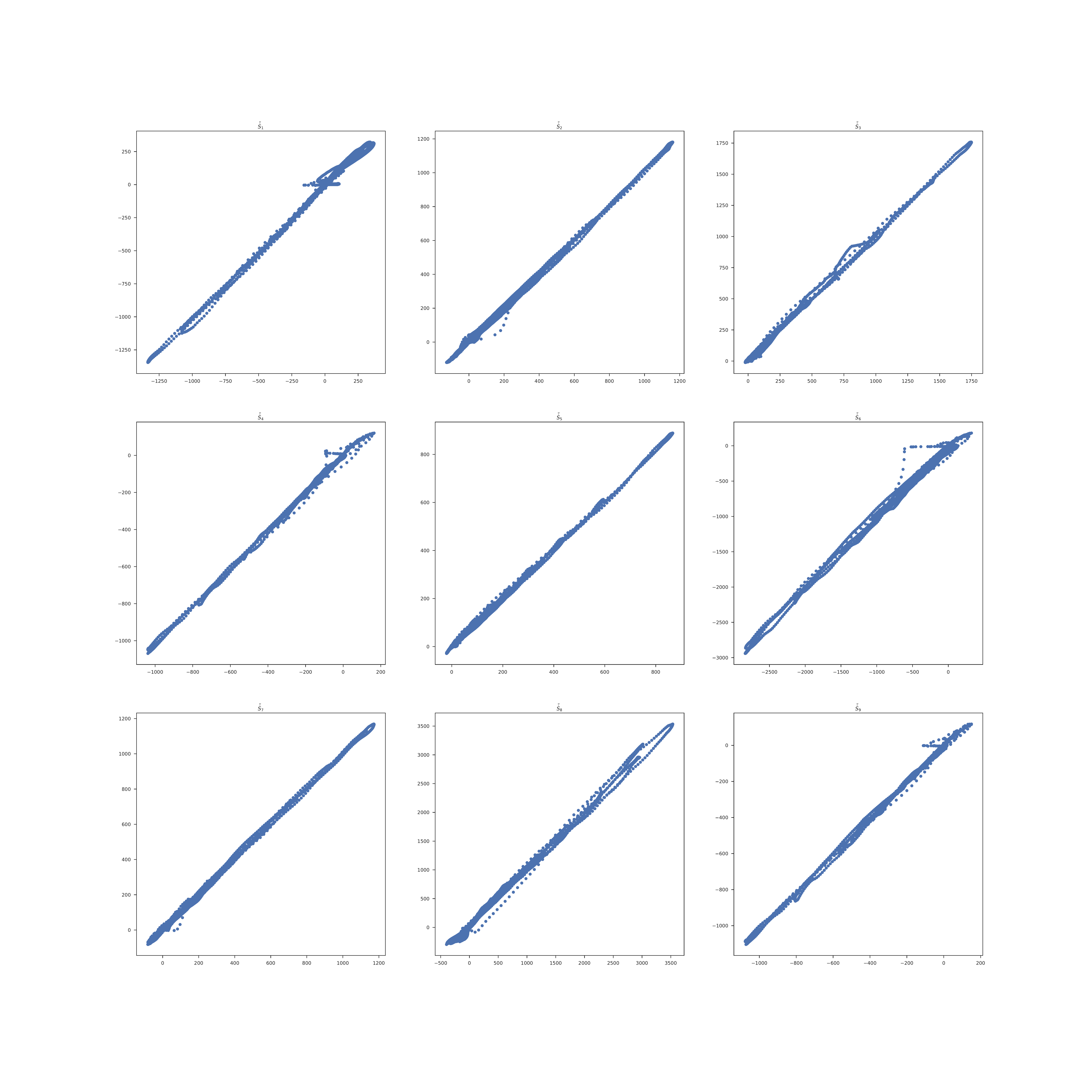}
  \caption{True Value vs Predicted Value Plots - Dynamic Source Term Regressor}
  \label{fig:qq-dynamic-source-prediction}
\end{figure}

In the figure \ref{fig:qq-dynamic-source-prediction} it can be observed that the {\it Dynamic Source Term Regressor} model does reasonably well and is able to capture the variability across the range adequately. We believe a post-training with a stratified oversampling and a combination of an ensemble for the {\it Dynamic Source Term Regressor} will improve the model performance even further.

\subsubsection{Residual Analysis}
In this section we present residual analysis of the {\it Physics Regressor}. 
\begin{figure}[H]
  \centering
  \includegraphics[width=0.4\linewidth]{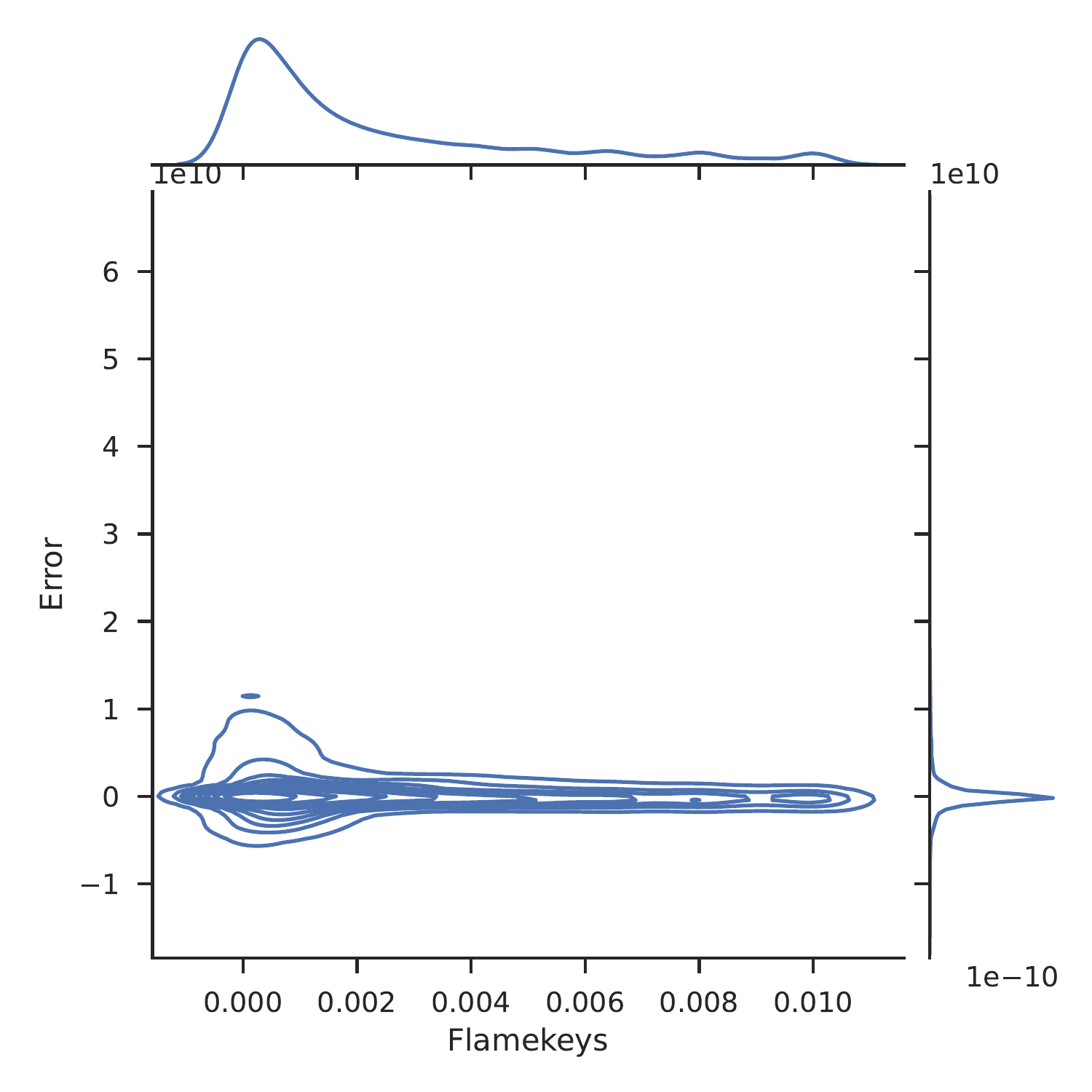}
  \includegraphics[width=0.4\linewidth]{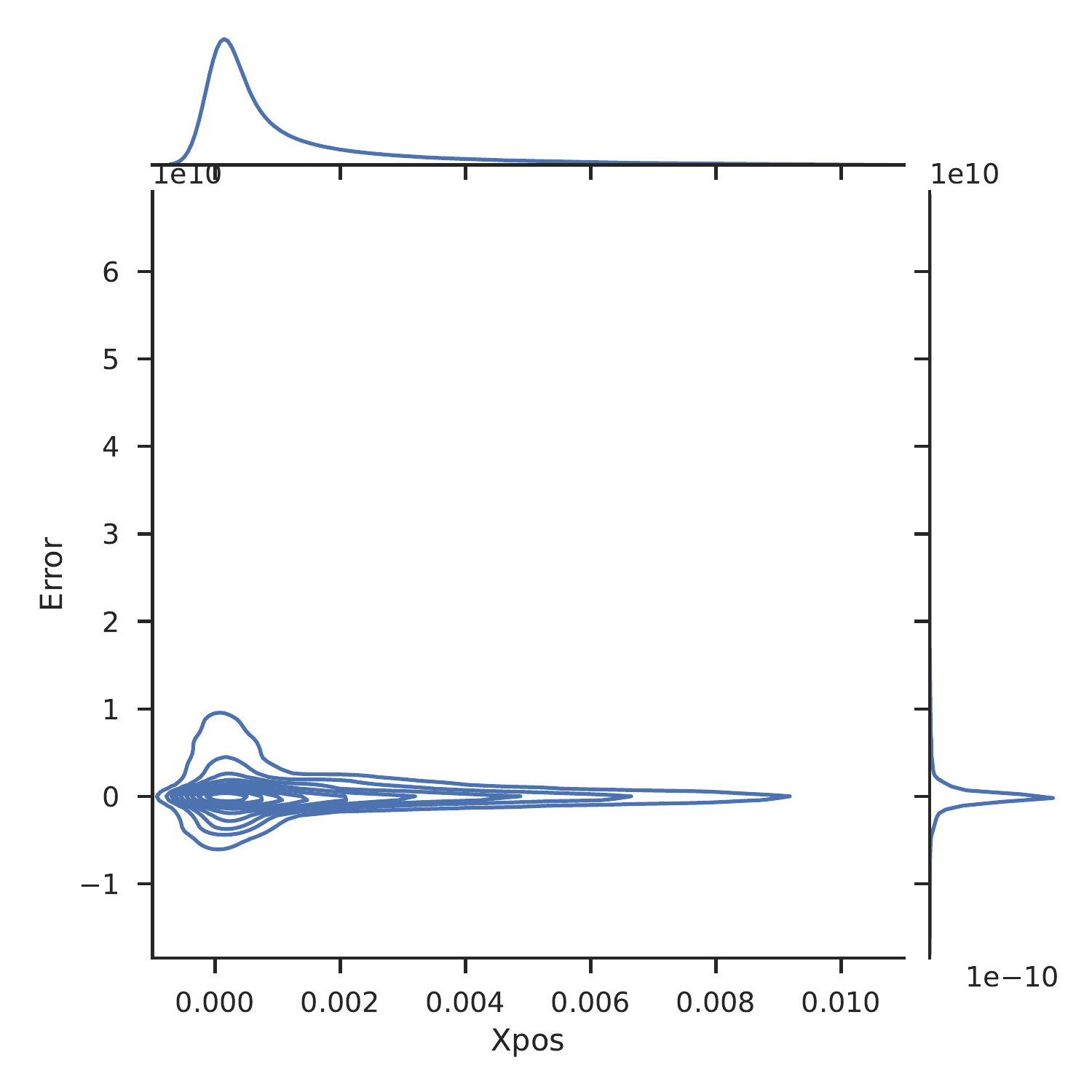}
  \includegraphics[width=0.4\linewidth]{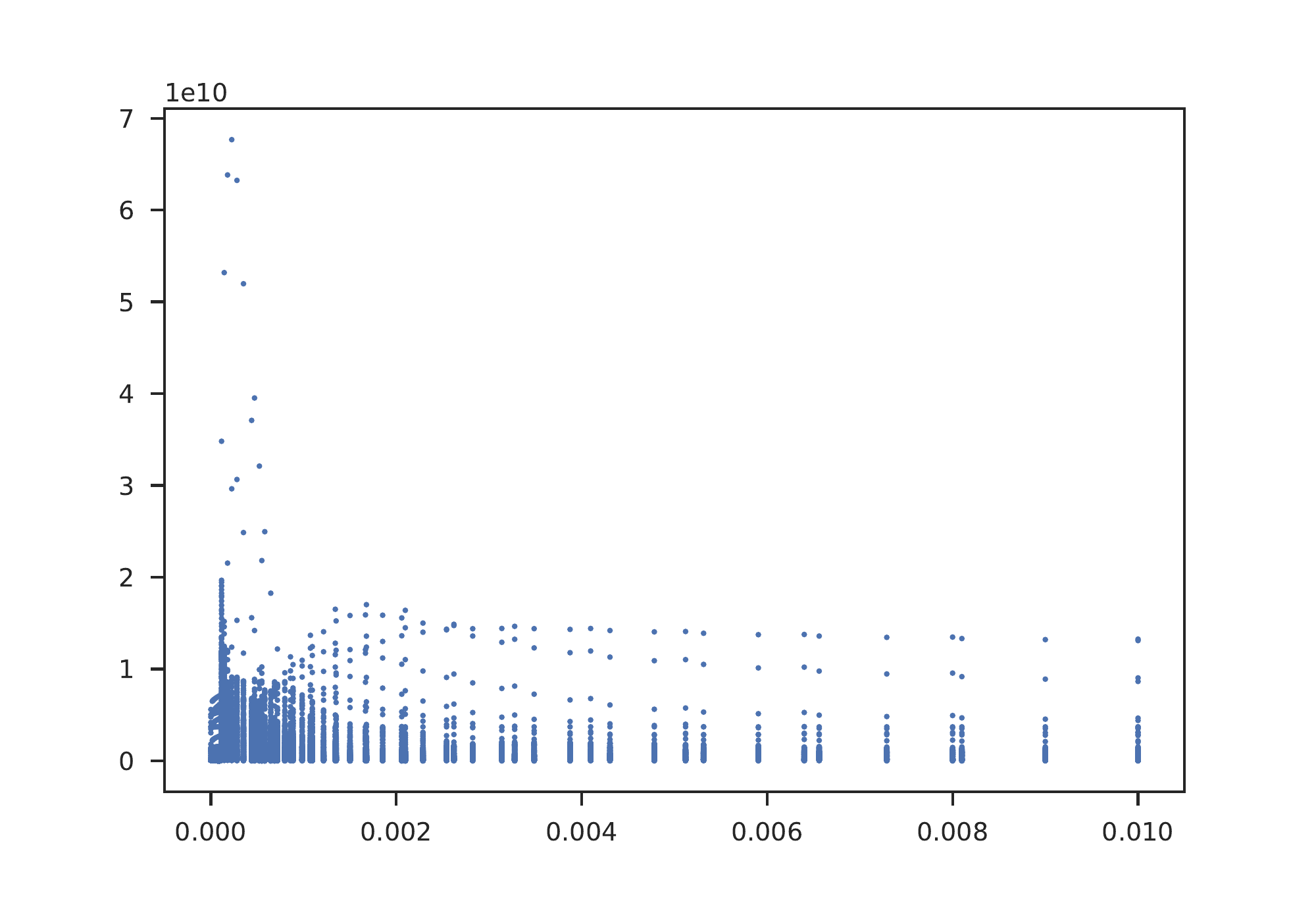}
    \includegraphics[width=0.4\linewidth]{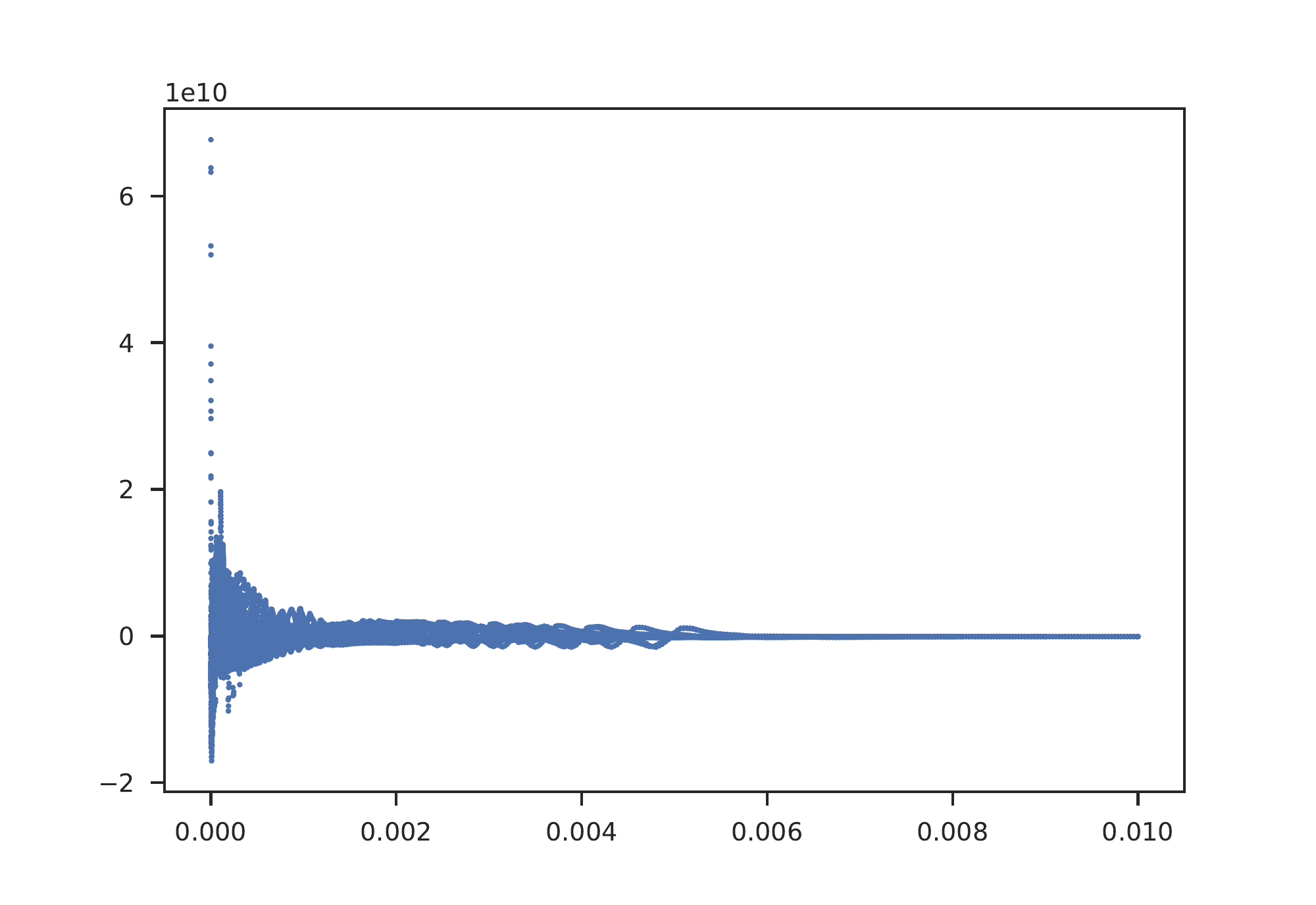}
  \caption{Source Energy Residuals by Flamekeys and Xpos}
  \label{fig:ext-residual-charts}
\end{figure}
In the figure \ref{fig:ext-residual-charts} we observe a concentration of residuals in the intial ranges of both the Flame Key and the Xpos. This is where the maximum combustion activity with high volatility exists. The problem could be simplifed by trying to learn this volatile behavior separately.  

\begin{figure}[H]
  \centering
  \includegraphics[width=1.0\linewidth]{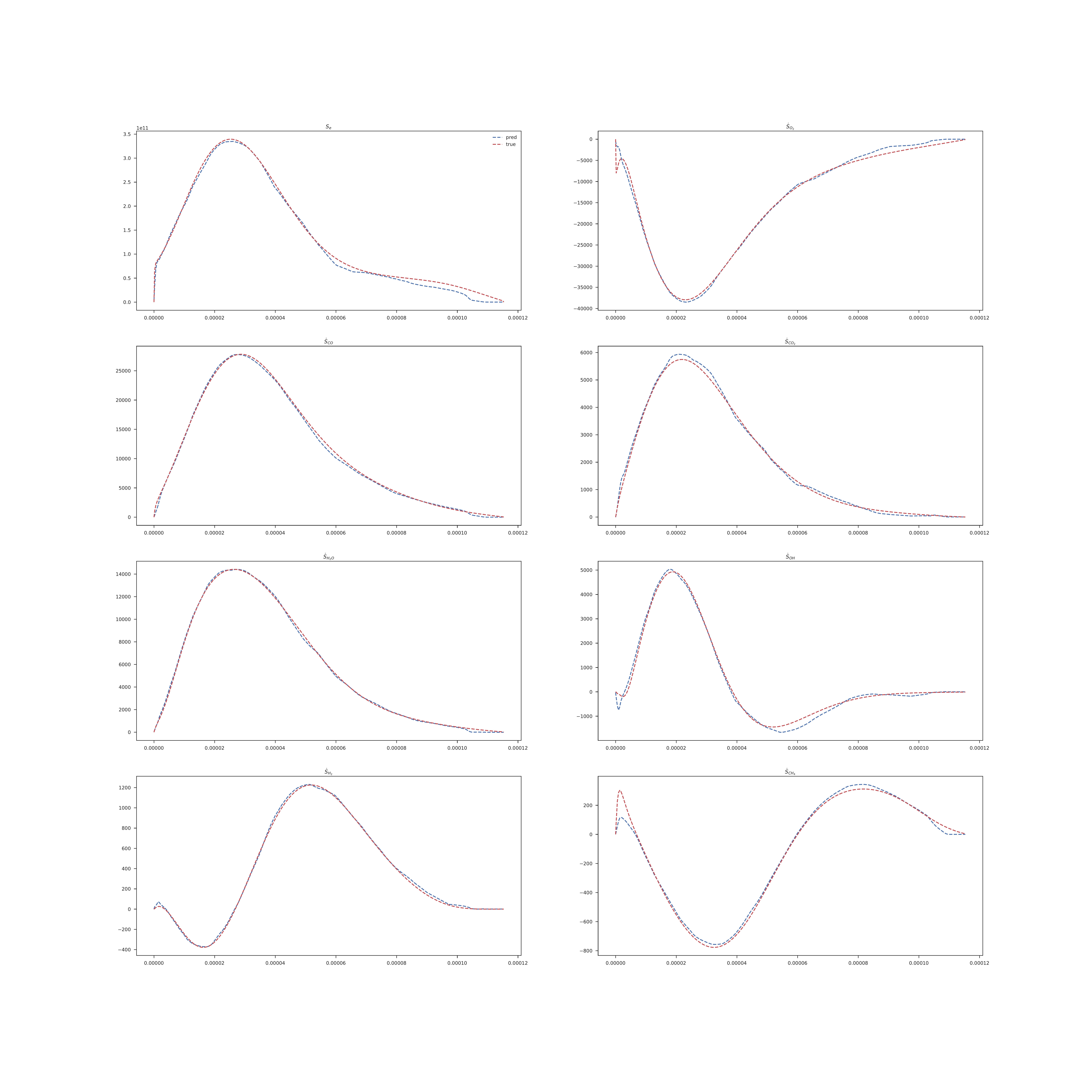}
  \caption{Model Performance for Flamelet Key = 0.00011529}
  \label{fig:extended-flamelet-high-activation}
\end{figure}

In the figure \ref{fig:extended-flamelet-high-activation} we present the {\it Physics Regressor} model prediction against the true values. The true data belongs to a particular flamelet and represents a highly active/combustive flame. The X-axis is the 'Xpos' and Y-axis is the value of the variable.

\begin{figure}[H]
  \centering
  \includegraphics[width=1.0\linewidth]{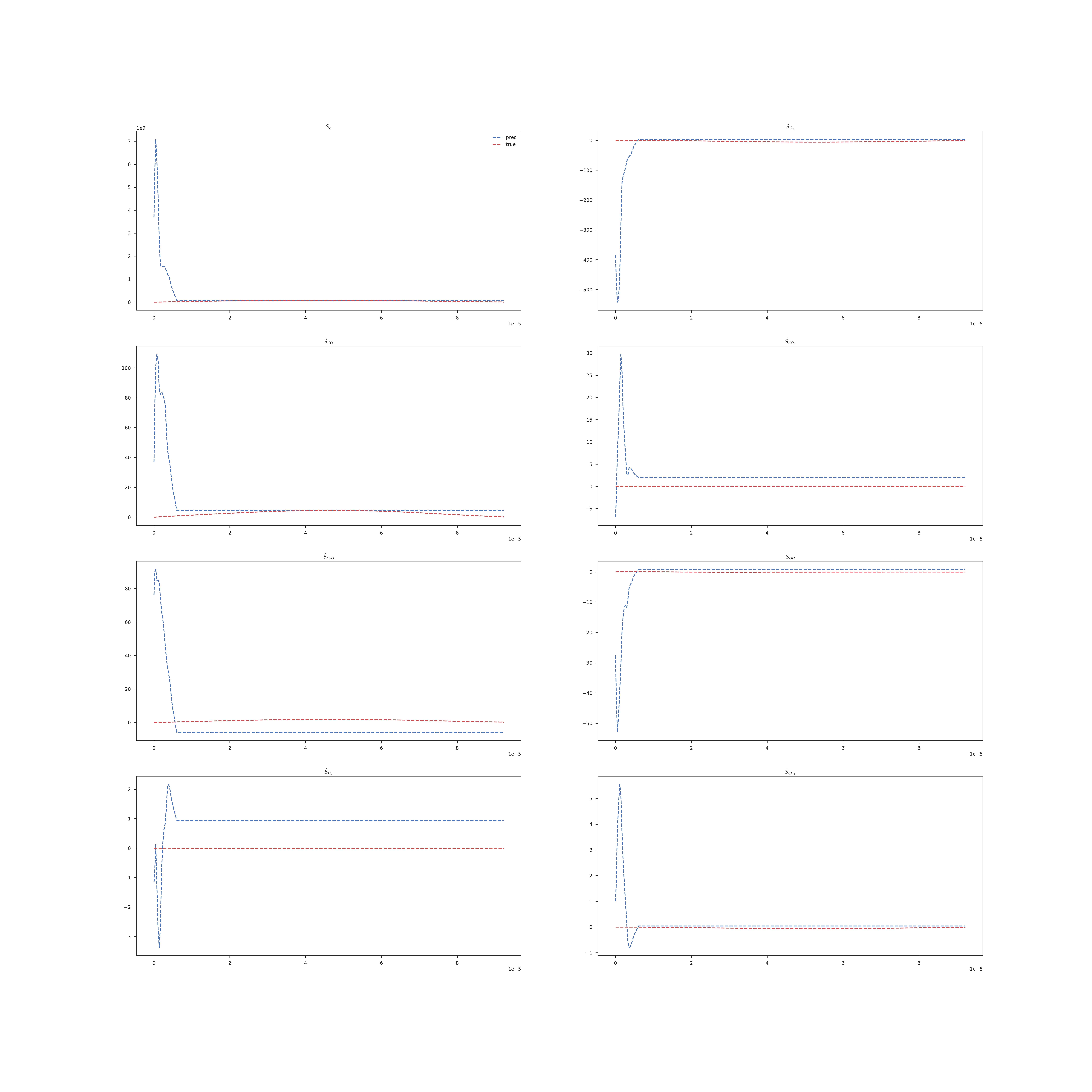}
  \caption{Model Performance for Flamelet Key = 0.00009223 }
  \label{fig:extended-flamelet-zero-activation}
\end{figure}

In the figure \ref{fig:extended-flamelet-zero-activation} the true data belongs to a particular flamelet and represents a non-active/extinguished flame. The model does a reasonable job and struggles in the initial ranges of the 'Xpos'. This is consistent with the observation in the \ref{fig:ext-residual-charts} residual charts.  We can certainly improve the model performance by adopting a stratified batch construction where each batch has a portion of the extinguished flames data. Alternatively, we can train a separate model for this data and/or create an ensemble to improve the overall performance.

\subsubsection{Constraint Satisfaction}
In this section we present the results on the constraints of the implementation.
\begin{table}[H]
\centering
  \caption{Non--Negative Constraint on the Weights of the Linear Embedding} 
  \small
        \begin{tabular}{|l|l|l|l|l|l|l|l|l|}
            \hline
             $w_{1}$ & $w_{2}$ & $w_{3}$ & $w_{4}$& $w_{5}$ & $w_{6}$ & $w_{7}$& $w_{8}$ & $w_{9}$\\
            \hline
          0.02 &  0.0   &   0.0   &  0.0   &   0.0   &  0.0   &   0.0   &  0.0   &  0.0  \\ 
       0.0   &  0.0   &  0.0   &  0.0   &  0.0   &  0.0   &  0.0   &   0.17 &  0.0  \\ 
        0.01 &   0.01 &  0.0   &   0.01 &   0.0   &  0.0   &   0.02 &   0.0   &   0.01\\ 
        0.02 &   0.0   &   0.0   &   0.03 &   0.0   &   0.05 &   0.0   &   0.0   &   0.03\\ 
        0.0   &  0.0   &   0.01 &   0.0   &  0.0   &   0.0   &  0.0   &   0.07 &   0.0  \\ 
        0.0   &   0.0   &   0.12 &   0.0   &   0.02 &   0.01 &   0.01 &   0.0   &   0.0  \\ 
        0.01 &   0.0   &  0.0   &   0.02 &  0.0   &  0.0   &  0.0   &  0.0   &   0.02\\ 
       0.0   &  0.0   &   0.31 &  0.0   &  0.0   &  0.0   &  0.0   &  0.0   &  0.0  \\ 
       0.0   &  0.0   &  0.0   &  0.0   &  0.0   &  0.0   &  0.0   &   0.48 &  0.0  \\ 
        0.4  &  0.0   &  0.0   &  0.0   &  0.0   &  0.0   &  0.0   &  0.0   &  0.0  \\ 
       0.0   &  0.0   &   0.66 &  0.0   &  0.0   &  0.0   &  0.0   &  0.0   &  0.0  \\ 
       0.0   &  0.0   &  0.0   &  0.0   &  0.0   &  0.0   &   0.33 &  0.0   &  0.0  \\ 
       0.0   &  0.0   &  0.0   &  0.0   &  0.0   &  0.0   &  0.0   &   0.24 &   0.0  \\ 
        0.42 &  0.0   &  0.0   &  0.0   &  0.0   &  0.0   &  0.0   &   0.0   &  0.0  \\ 
        0.0   &   0.04 &  0.0   &   0.0   &   0.02 &   0.0   &   0.04 &   0.05 &   0.0  \\ 
        0.0   &  0.0   &   0.0   &   0.0   &   0.0   &   0.0   &  0.0   &   0.24 &   0.0  \\ 
       0.0   &  0.0   &  0.0   &  0.0   &   0.08 &  0.0   &  0.0   &  0.0   &  0.0  \\ 
       0.0   &  0.0   &   0.0   &  0.0   &  0.0   &  0.0   &  0.0   &  0.0   &   0.11\\ 
       0.0   &  0.0   &  0.0   &  0.0   &  0.0   &   0.58 &  0.0   &  0.0   &  0.0  \\ 
        0.64 &  0.0   &  0.0   &  0.0   &  0.0   &  0.0   &  0.0   &  0.0   &  0.0  \\ 
       0.0   &   0.39 &  0.0   &  0.0   &  0.0   &  0.0   &  0.0   &  0.0   &  0.0  \\ 
       0.0   &   0.1  &  0.0   &  0.0   &  0.0   &  0.0   &  0.0   &  0.0   &  0.0  \\ 
       0.0   &   0.01 &   0.0   &  0.0   &   0.0   &   0.1  &  0.0   &  0.0   &  0.0  \\ 
       0.0   &  0.0   &  0.0   &   0.0   &  0.0   &   0.0   &   0.05 &  0.0   &   0.0  \\ 
        0.01 &  0.0   &   0.0   &  0.0   &   0.0   &   0.1  &   0.0   &  0.0   &  0.0  \\ 
       0.0   &  0.0   &  0.0   &   0.16 &  0.0   &  0.0   &  0.0   &  0.0   &  0.0  \\ 
       0.0   &  0.0   &  0.0   &  0.0   &  0.0   &   0.54 &  0.0   &  0.0   &  0.0  \\ 
       0.0   &  0.0   &  0.0   &  0.0   &  0.0   &   0.28 &  0.0   &  0.0   &  0.0  \\ 
       0.0   &  0.0   &  0.0   &  0.0   &  0.0   &   0.14 &  0.0   &  0.0   &  0.0  \\ 
       0.0   &  0.0   &  0.0   &  0.0   &   0.3  &  0.0   &  0.0   &  0.0   &  0.0  \\ 
       0.0   &  0.0   &   0.57 &  0.0   &  0.0   &  0.0   &  0.0   &  0.0   &  0.0  \\ 
       0.0   &  0.0   &  0.0   &  0.0   &  0.0   &  0.0   &  0.0   &  0.0   &   0.62\\ 
       0.0   &   0.5  &  0.0   &  0.0   &  0.0   &  0.0   &  0.0   &  0.0   &  0.0  \\ 
       0.0   &   0.53 &  0.0   &  0.0   &  0.0   &  0.0   &  0.0   &  0.0   &  0.0  \\ 
       0.0   &  0.0   &  0.0   &  0.0   &  0.0   &   0.5  &  0.0   &  0.0   &  0.0  \\ 
       0.0   &  0.0   &  0.0   &   0.42 &  0.0   &  0.0   &  0.0   &  0.0   &  0.0  \\ 
       0.0   &  0.0   &  0.0   &   0.32 &  0.0   &  0.0   &  0.0   &  0.0   &  0.0  \\ 
       0.0   &  0.0   &  0.0   &  0.0   &   0.42 &  0.0   &  0.0   &  0.0   &  0.0  \\ 
       0.0   &  0.0   &  0.0   &  0.0   &   0.5  &  0.0   &  0.0   &  0.0   &  0.0  \\ 
       0.0   &  0.0   &  0.0   &  0.0   &  0.0   &  0.0   &   0.74 &  0.0   &  0.0  \\ 
       0.0   &  0.0   &  0.0   &  0.0   &  0.0   &  0.0   &  0.0   &  0.0   &   0.62\\ 
       0.0   &  0.0   &  0.0   &   0.44 &  0.0   &  0.0   &  0.0   &  0.0   &  0.0  \\ 
       0.0   &  0.0   &  0.0   &   0.52 &  0.0   &  0.0   &  0.0   &  0.0   &  0.0  \\ 
       0.0   &  0.0   &   0.36 &  0.0   &  0.0   &  0.0   &  0.0   &  0.0   &  0.0  \\ 
       0.0   &  0.0   &  0.0   &  0.0   &   0.53 &  0.0   &  0.0   &  0.0   &  0.0  \\ 
       0.0   &  0.0   &  0.0   &  0.0   &   0.45 &  0.0   &  0.0   &  0.0   &  0.0  \\ 
       0.0   &  0.0   &  0.0   &   0.49 &  0.0   &  0.0   &  0.0   &  0.0   &  0.0  \\ 
        0.5  &  0.0   &  0.0   &  0.0   &  0.0   &  0.0   &  0.0   &  0.0   &  0.0  \\ 
       0.0   &  0.0   &  0.0   &  0.0   &  0.0   &  0.0   &  0.0   &   0.79 &  0.0  \\ 
       0.0   &  0.0   &  0.0   &  0.0   &  0.0   &  0.0   &   0.03 &  0.0   &  0.0  \\ 
       0.0   &  0.0   &  0.0   &  0.0   &  0.0   &  0.0   &   0.58 &  0.0   &  0.0  \\ 
       0.0   &  0.0   &  0.0   &  0.0   &  0.0   &  0.0   &  0.0   &  0.0   &   0.47\\ 
       0.0   &   0.55 &  0.0   &  0.0   &  0.0   &  0.0   &  0.0   &  0.0   &  0.0  \\
           \hline
        \end{tabular}
        \label{tab:extended-nn-constraints}
\end{table}

As can be observed in table \ref{tab:extended-nn-constraints} all the weights are non-negative and the \ref{eqn:ext-NN-constraint} constraint is completely satisfied.

\begin{table}[H]
    \centering
    \caption{ Orthogonality Constraint on the Weights of the\\
    Linear Embedding}
\begin{tabular}{|l|l|l|l|l|l|l|l|l|l|}
    \hline
      & $w_{1}$ & $w_{2}$ & $w_{3}$ & $w_{4}$& $w_{5}$ & $w_{6}$ & $w_{7}$& $w_{8}$ & $w_{9}$\\
    \hline
 	$w_{1}$ & 0.997& 0.0  & 0.0  & 0.001& 0.0  & 0.002& 0.0  & 0.0  & 0.001\\
    $w_{2}$ & 0.0  & 0.997& 0.0  & 0.0  & 0.001& 0.001& 0.002& 0.002& 0.0  \\
    $w_{3}$ & 0.0  & 0.0  & 1.001& 0.0  & 0.002& 0.001& 0.001& 0.001& 0.0  \\
    $w_{4}$ & 0.001& 0.0  & 0.0  & 1.01 & 0.0  & 0.002& 0.0  & 0.0  & 0.001\\
    $w_{5}$ & 0.0  & 0.001& 0.002& 0.0  & 1.007& 0.0  & 0.001& 0.001& 0.0  \\
    $w_{6}$ & 0.002& 0.001& 0.001& 0.002& 0.0  & 0.999& 0.0  & 0.0  & 0.002\\
    $w_{7}$ & 0.0  & 0.002& 0.001& 0.0  & 0.001& 0.0  & 0.998& 0.002& 0.0  \\
    $w_{8}$ & 0.0  & 0.002& 0.001& 0.0  & 0.001& 0.0  & 0.002& 1.006& 0.0  \\
    $w_{9}$ & 0.001& 0.0  & 0.0  & 0.001& 0.0  & 0.002& 0.0  & 0.0  & 1.003\\
   \hline
\end{tabular}
\label{tab:extended-wo-constraints}
\end{table}
The \ref{eqn:ext-WO-constraint} constraint conformity is measured through covariance of the output of the linear encoder. As can be observed in table \ref{tab:extended-wo-constraints} all the diagonal entries are close to 1 and the non-diagonal entries are close to 0 indicating that the Orthogonality constraint on the weights is satisfied.

\begin{table}[H]
    \centering
    \caption{Orthogonality Constraint on the Output of the\\
       Linear Embedding}
     \begin{tabular}{|l|l|l|l|l|l|l|l|l|l|l|}
        \hline
          & $Z_{mix}$& $C_{pv_{1}}$ & $C_{pv_{2}}$ & $C_{pv_{3}}$ & $C_{pv_{4}}$& $C_{pv_{5}}$ & $C_{pv_{6}}$ & $C_{pv_{7}}$ & $C_{pv_{8}}$& $C_{pv_{9}}$\\
        \hline
        $Z_{mix}$ & 0.0 &    0.0 &   0.0 &    0.0 &   0.0 &   0.0 &    0.0 &   0.0 &    0.0 &   0.0  \\
         $C_{pv_{1}}$ & 0.0 &    0.0 &    0.0 &   0.0 &    0.0 &   0.0 &    0.0 &    0.0 &   0.0 &    0.0  \\
         $C_{pv_{2}}$ &0.0 &    0.0 &    0.0 &   0.0 &    0.0 &   0.0 &    0.0 &    0.0 &   0.0 &   -0.001 \\
         $C_{pv_{3}}$ & 0.0 &   0.0 &   0.0 &    0.0 &   0.0 &    0.0 &   0.0 &   0.0 &    0.0 &   -0.001 \\
         $C_{pv_{4}}$ &0.0 &    0.0 &    0.0 &   0.0 &    0.0 &   0.0 &    0.0 &    0.0 &   0.0 &   0.0  \\
         $C_{pv_{5}}$ &0.0 &   0.0 &   0.0 &    0.0 &   0.0 &    0.001 & 0.0 &   -0.001  &0.0 &    0.005 \\
         $C_{pv_{6}}$ & 0.0 &    0.0 &    0.0 &   0.0 &    0.0 &   0.0 &    0.0 &    0.0 &   0.0 &    0.0  \\
         $C_{pv_{7}}$ &0.0 &    0.0 &    0.0 &   0.0 &    0.0 &   -0.001 &  0.0 &    0.001 & 0.0 &   -0.002 \\
         $C_{pv_{8}}$ & 0.0 &   0.0 &   0.0 &    0.0 &   0.0 &    0.0 &   0.0 &   0.0 &    0.0 &   -0.001 \\
         $C_{pv_{9}}$ &0.0 &    0.0 &   -0.001 & -0.001 & 0.0 &    0.005 &  0.0 &   -0.002 &-0.001  &0.067 \\
       \hline
    \end{tabular} 
    \label{tab:extended-ar-constraints}
\end{table}
The \ref{eqn:ext-AR-constraint} constraint conformity is measured through covariance of the output of the linear encoder. As can be observed in table \ref{tab:extended-ar-constraints} 
all the non-diagonal entries are close to 0 indicating that the Orthogonality constraint on the $C_{pv}$ is satisfied. 

\section{Conclusion}
Building on our prior work \cite{ChemTab} we presented an extended formulation for jointly learning the progress variables and the manifold approximation for solving the high--dimensional chemistry in combustion models with newer constraints and a novel {\it Dynamic Source Term Regressor} and showcased the generalizability of this approach across multiple datasets. Our approach follows the principle of physics guided neural networks~\cite{karpatne2017}, which are increasingly becoming popular for many scientific modeling tasks, though no solutions exist that can directly benefit the combustion community. Our formulation outperforms the state--of--the--art state--space parametrization in combustion. The generated reaction progress variables can be interpreted by examining the projection/weight matrix, {\em $\mathcal{W}$}, and thus, allows for physical insights into the systems being modeled. This formulation leverages the projection matrix to dynamically create additional thermochemical state variables which are learnt by the {\it Dynamic Source Term Regressor}r--the ease/difficulty of this learning task in turn influences the projection matrix.  

In the future we will work on extending the current formulation with an Autoencoder of the mass fraction which will allow us to incorporate the influence of reconstruction error in the learning of the embedding. Quantification of uncertainity in the estimation of the Thermochemical State variables will also be our next focus. We believe the deep neural network based implementation lends naturally to the adoption of deep ensembles for uncertainty quantification.

\section{Acknowledgments}
Funded by the United States Department of Energy’s (DoE) National Nuclear Security Administration (NNSA) under the Predictive Science Academic Alliance Program III (PSAAP III) at the University at Buffalo, under contract number DE--NA0003961.

\bibliographystyle{splncs04}

\begin{thebibliography}{10}
\providecommand{\url}[1]{\texttt{#1}}
\providecommand{\urlprefix}{URL }
\providecommand{\doi}[1]{https://doi.org/#1}

\bibitem{bhalla2019}
Bhalla, S., Yao, M., Hickey, J.P., Crowley, M.: Compact representation of a
  multi-dimensional combustion manifold using deep neural networks. In:
  European Conference on Machine Learning (2019)

\bibitem{biglari20154025}
Biglari, A., Sutherland, J.C.: An a-posteriori evaluation of principal
  component analysis-based models for turbulent combustion simulations.
  Combustion and Flame  \textbf{162}(10),  4025--4035 (2015)

\bibitem{bojko2016}
Bojko, B.T., DesJardin, P.E.: Formulation and assessment of flamelet-generated
  manifolds for reacting interfaces. Combustion and Flame  \textbf{173},
  296--306 (2016)

\bibitem{chen2011}
Chen, J.: Petascale direct numerical simulation of turbulent combustion -
  fundamental insights towards predictive models. Proceedings of the Combustion
  Institute  \textbf{33},  99–123 (12 2011).
  \doi{10.1016/j.proci.2010.09.012}

\bibitem{ElAsrag2013ACB}
El-Asrag, H.A.: A comparison between two different flamelet reduced order
  manifolds for non-premixed turbulent flames. In: Laminar \& Turbulent Flames
  (2013)

\bibitem{gicquel2004}
Fiorina, B., Gicquel, O., Carpentier, S., Darabiha, N.: Validation of the fpi
  chemistry reduction method for diluted nonadiabatic premixed flames.
  Combustion Science and Technology  \textbf{176}(5-6),  785--797 (2004)

\bibitem{Giusti:2019}
Giusti, A., Mastorakos, E.: Turbulent combustion modelling and experiments:
  Recent trends and developments. Flow, Turbulence and Combustion
  \textbf{103}(4),  847--869 (2019)

\bibitem{ihme2005}
Ihme, M., Cha, C., Pitsch, H.: Prediction of local extinction and re-ignition
  effects in non-premixed turbulent combustion using a flamelet/progress
  variable approach. Proceedings of the Combustion Institute  \textbf{30},
  793--800 (01 2005)

\bibitem{ihme2009}
Ihme, M., Schmitt, C., Pitsch, H.: Optimal artificial neural networks and
  tabulation methods for chemistry representation in les of a bluff-body
  swirl-stabilized flame. Proceedings of the Combustion Institute  \textbf{32},
   1527--1535 (12 2009)

\bibitem{ihme20127715}
Ihme, M., Shunn, L., Zhang, J.: Regularization of reaction progress variable
  for application to flamelet-based combustion models. Journal of Computational
  Physics  \textbf{231}(23),  7715--7721 (2012)

\bibitem{karpatne2017}
Karpatne, A., Atluri, G., Faghmous, J.H., Steinbach, M., Banerjee, A., Ganguly,
  A., Shekhar, S., Samatova, N., Kumar, V.: Theory-guided data science: A new
  paradigm for scientific discovery from data. IEEE Transactions on Knowledge
  and Data Engineering  \textbf{29}(10),  2318--2331 (2017)

\bibitem{lu2009}
Lu, T., Law, C.K.: Toward accommodating realistic fuel chemistry in large-scale
  computations. Progress in Energy and Combustion Science  \textbf{35}(2),
  192--215 (2009). \doi{https://doi.org/10.1016/j.pecs.2008.10.002},
  \url{https://www.sciencedirect.com/science/article/pii/S036012850800066X}

\bibitem{maas1992}
Maas, U., Pope, S.B.: Implementation of simplified chemical kinetics based on
  intrinsic low-dimensional manifolds. In: Symposium (International) on
  Combustion. vol.~24, pp. 103--112. Elsevier (1992)

\bibitem{malik201830}
Malik, M.R., Isaac, B.J., Coussement, A., Smith, P.J., Parente, A.: Principal
  component analysis coupled with nonlinear regression for chemistry reduction.
  Combustion and Flame  \textbf{187},  30--41 (2018)

\bibitem{malik2020}
Malik, M.R., {Obando Vega}, P., Coussement, A., Parente, A.: Combustion
  modeling using principal component analysis: A posteriori validation on
  sandia flames {D}, {E} and {F}. Proceedings of the Combustion Institute
  (2020)

\bibitem{montgomery2005}
Montgomery, M.J., Kwon, H., Kastengren, A.L., Pfefferle, L.D., Sikes, T.,
  Tranter, R.S., Xuan, Y., McEnally, C.S.: In situ temperature measurements in
  sooting methane/air flames using synchrotron x-ray fluorescence of seeded
  krypton atoms. Science Advances  \textbf{8}(17),  eabm7947 (2022).
  \doi{10.1126/sciadv.abm7947},
  \url{https://www.science.org/doi/abs/10.1126/sciadv.abm7947}

\bibitem{nouri2019}
Nouri, A.G., Givi, P., Livescu, D.: Modeling and simulation of turbulent
  nuclear flames in type ia supernovae. Progress in Aerospace Sciences
  \textbf{108},  156--179 (2019).
  \doi{https://doi.org/10.1016/j.paerosci.2019.04.004},
  \url{https://www.sciencedirect.com/science/article/pii/S037604211930020X}

\bibitem{Yellapantula2021}
Perry, B.A., de~Frahan, M.T.H., Yellapantula, S.: Evaluation of co-optimized
  machine-learned manifolds for modeling premixed combustion (2021),
  \url{https://ui.adsabs.harvard.edu/abs/2021APS..DFDF09003P/abstract}

\bibitem{PETERS1984319}
Peters, N.: Laminar diffusion flamelet models in non-premixed turbulent
  combustion. Progress in Energy and Combustion Science  \textbf{10}(3),
  319--339 (1984)

\bibitem{peters2001}
Peters, N.: Turbulent combustion. IOP Publishing (2001)

\bibitem{piercemoin2004}
Pierce, C.D., Moin, P.: Progress-variable approach for large-eddy simulation of
  non-premixed turbulent combustion. Journal of Fluid Mechanics  \textbf{504},
  73–97 (2004)

\bibitem{raisi2018}
Raissi, M., Karniadakis, G.E.: Hidden physics models: Machine learning of
  nonlinear partial differential equations. Journal of Computational Physics
  \textbf{357},  125--141 (2018).
  \doi{https://doi.org/10.1016/j.jcp.2017.11.039},
  \url{https://www.sciencedirect.com/science/article/pii/S0021999117309014}

\bibitem{raissi2020}
Raissi, M., Yazdani, A., Karniadakis, G.E.: Hidden fluid mechanics: Learning
  velocity and pressure fields from flow visualizations. Science
  \textbf{367}(6481),  1026--1030 (2020). \doi{10.1126/science.aaw4741},
  \url{https://www.science.org/doi/abs/10.1126/science.aaw4741}

\bibitem{Rastigejev13875}
Rastigejev, Y., Brenner, M.P., Jacob, D.J.: Spatial reduction algorithm for
  atmospheric chemical transport models. Proceedings of the National Academy of
  Sciences  \textbf{104}(35),  13875--13880 (2007)

\bibitem{ChemTab}
Salunkhe, A., Deighan, D., DesJardin, P.E., Chandola, V.: Chemtab: A physics
  guided chemistry modeling framework. In: Groen, D., de~Mulatier, C.,
  Paszynski, M., Krzhizhanovskaya, V.V., Dongarra, J.J., Sloot, P.M.A. (eds.)
  Computational Science -- ICCS 2022. pp. 75--88. Springer International
  Publishing, Cham (2022)

\bibitem{grimech}
Smith, G.P., Golden, D.M., Frenklach, M., Moriarty, N.W., Eiteneer, B.,
  Goldenberg, M., Bowman, C.T., Hanson, R.K., Song, S., Gardiner~Jr., W.C.,
  Lissianski, V.V., Qin, Z.: Gri-mech 3.0 is an optimized mechanism designed to
  model natural gas combustion, including no formation and reburn chemistry.

\bibitem{sutherland20091563}
Sutherland, J.C., Parente, A.: Combustion modeling using principal component
  analysis. Proceedings of the Combustion Institute  \textbf{32}(1),
  1563--1570 (2009)

\bibitem{fgm2000}
Van~Oijen, J., De~Goey, L.: Modelling of premixed laminar flames using
  flamelet-generated manifolds. Combustion Science and Technology
  \textbf{161}(1),  113--137 (2000)

\bibitem{van2001}
{van Oijen}, J., Lammers, F., {de Goey}, L.: Modeling of complex premixed
  burner systems by using flamelet-generated manifolds. Combustion and Flame
  \textbf{127}(3),  2124--2134 (2001)

\bibitem{willard2020}
Willard, J., Jia, X., Xu, S., Steinbach, M., Kumar, V.: Integrating scientific
  knowledge with machine learning for engineering and environmental systems
  (2021)

\bibitem{zhang20201}
Zhang, Y., Xu, S., Zhong, S., Bai, X.S., Wang, H., Yao, M.: Large eddy
  simulation of spray combustion using flamelet generated manifolds combined
  with artificial neural networks. Energy and AI  \textbf{2},  100021 (2020)

\end{thebibliography}

\end{document}